\documentclass[11pt]{article}
\usepackage{acl}

\usepackage{times}
\usepackage{caption}
\usepackage{latexsym}
\usepackage[T1]{fontenc}
\usepackage[utf8]{inputenc}
\usepackage{microtype}
\usepackage{inconsolata}
\usepackage{booktabs}
\usepackage{amsmath,amssymb}
\usepackage{graphicx}
\usepackage{dblfloatfix}
\usepackage{xcolor}
\usepackage{multirow}
\usepackage{tabularx}
\usepackage{adjustbox}
\usepackage{enumitem}
\usepackage{placeins}
\usepackage{float}
\usepackage{url}
\usepackage{xspace}

\usepackage[most]{tcolorbox}
\usepackage{fvextra}

\definecolor{promptframe}{RGB}{150,155,165}
\definecolor{promptback}{RGB}{250,250,251}
\definecolor{prompttitle}{RGB}{239,241,244}

\newtcolorbox{bcareprompt}[1]{
  enhanced,
  breakable,
  colback=promptback,
  colframe=promptframe,
  colbacktitle=prompttitle,
  coltitle=black,
  fonttitle=\bfseries\footnotesize,
  title={#1},
  boxrule=0.45pt,
  arc=1.5pt,
  outer arc=1.5pt,
  left=6pt,
  right=6pt,
  top=6pt,
  bottom=6pt,
  before skip=7pt,
  after skip=9pt,
  fontupper=\footnotesize
}

\newcommand{\promptheading}[1]{%
  \par\smallskip\noindent\textbf{#1}\par\smallskip
}
\newcommand{\promptfield}[1]{%
  \texttt{\detokenize{#1}}%
}

\title{B-CARE: Blind Classwise Assessment of Relative Evidence for Stance Error Ranking}
\author{
  \textbf{Meysam Sabbaghan \quad Doina Caragea} \\
  Department of Computer Science \\
  Kansas State University \\
  \texttt{\{meysamsabbaghan,dcaragea\}@ksu.edu}
}
\date{}
\begin{document}
\maketitle

\begin{abstract}
Stance classifiers make errors that can be costly to identify
under limited human-review budgets. Prioritizing potentially
incorrect predictions remains challenging, particularly across
heterogeneous base models. We introduce
B-CARE (\textbf{B}lind \textbf{C}lasswise
\textbf{A}ssessment of \textbf{R}elative
\textbf{E}vidence for Stance Error Ranking),
a prediction-blind framework for identifying high-risk
stance predictions. Given only a target and text, a fixed
LLM verifier assigns ordinal evidence scores to all
candidate stance labels without observing the base model's
prediction or confidence. B-CARE constructs a comparative
risk score by contrasting the strongest alternative-class
evidence with the evidence supporting the predicted class.
The resulting classwise evidence profile can be reused
across multiple base models without training a separate
error detector or calibrating risk scores.
We evaluate B-CARE on four stance benchmarks and three
heterogeneous base LLMs, covering 12 dataset--model
settings and 5{,}206 unique test instances. Evaluation
includes error-ranking discrimination (AUROC and AUPRC),
selective prediction (E-AURC), and review effectiveness
(ErrorCapture at fixed review budgets). B-CARE achieves
a mean AUROC of 0.735 and a mean AUPRC of 0.524,
capturing 41.06\% of errors within the highest-risk
20\% of predictions. Although B-CARE does not uniformly
outperform strong confidence-based baselines, it provides
a reusable, complementary evidence signal for prioritizing
potentially incorrect stance predictions across
heterogeneous models.
The code is available in an anonymized repository.\footnote{\url{https://anonymous.4open.science/r/B-CARE-Code-EC84/}}
\end{abstract}

\section{Introduction}
\label{sec:introduction}

Stance detection identifies whether a text favors,
opposes, or takes no clear position toward a target
\cite{mohammad-etal-2016-semeval}. Despite advances
in LLM-based prompting \cite{akash-etal-2025-large} and reasoning \cite{zhang2026cognitive}, as well as extensive study of external-context integration \cite{nguyen-kim-2025-external},
classification errors persist. Under limited
human-review budgets, a key question is:
\emph{which predictions should be reviewed first?}

Existing reliability signals include maximum
softmax probability (MSP) \cite{hendrycks2017baseline},
predictive entropy (PE) \cite{malinin2018predictive},
and $p(\mathrm{True})$ \cite{kadavath2022language}.
Other approaches use semantic uncertainty or
similarity-based signals
\cite{kuhn2023semantic,farquhar2024detecting,
nguyen-etal-2025-beyond},
evidential uncertainty
\cite{kunitomo-jacquin-etal-2026-evidential},
and consistency- or verification-based checks
\cite{zhang-etal-2023-sac3,xue2025verify,
tan-etal-2025-consistent,goel-etal-2025-zero,
hamidieh2026crossmodel}.
Closest to our setting, MPVStance
\cite{zhang-etal-2025-mpvstance}
verifies and revises stance predictions, while
MARCH \cite{li-etal-2026-march} checks claims
without revealing the solver's answer to its
checker. B-CARE instead ranks errors in fixed
predictions using prediction-blind, graded
evidence for competing classes.

We introduce B-CARE
(\textbf{B}lind \textbf{C}lasswise
\textbf{A}ssessment of \textbf{R}elative
\textbf{E}vidence for Stance Error Ranking).
A fixed LLM verifier independently assesses
ordinal evidence $E_c(x)$ for each candidate
stance class without observing the base
prediction or confidence. For a base classifier
$B$ predicting $\hat{y}_B$, the error-risk score is
\begin{equation}
R_B(x)=
\max_{c\neq\hat{y}_B}E_c(x)-E_{\hat{y}_B}(x).
\end{equation}
Unlike binary verifier disagreement, this score
preserves the strength of predicted-class support
relative to competing evidence. The resulting
evidence profiles can be reused across classifiers
without training an error detector or performing
dataset-specific risk calibration.

We evaluate B-CARE on SemEval-2016 Task 6A
\cite{mohammad-etal-2016-semeval},
VaccineStance (1,000 examples)
\cite{sabbaghan2026moprosd},
COVID-19-Stance \cite{glandt-etal-2021-stance},
and P-Stance \cite{li-etal-2021-p}, with
Phi-4 \cite{abdin2024phi4},
Mistral Small 24B Instruct (2501)
\cite{mistralai2025small2501}, and
Gemma 3 \cite{gemmateam2025gemma3}.
Across 12 dataset--model settings and 5{,}206
unique test instances, B-CARE achieves mean
AUROC 0.735 and AUPRC 0.524, capturing 41.06\%
of existing errors at a 20\% review budget.
Ablations support prediction blindness and
graded evidence, although performance varies
across datasets and does not consistently exceed
strong confidence baselines. Exploratory results
also indicate complementarity with predictive
entropy. 
Our contributions are:
\begin{itemize}[leftmargin=*,nosep]
\item \textbf{Comparative error ranking:}
Prediction-blind classwise evidence and relative
scoring without a trained error detector.
\item \textbf{Controlled analysis:}
Ablations of prediction exposure, evidence
granularity, and score components.
\item \textbf{Evaluation and complementarity:}
Error ranking, selective prediction, and
review effectiveness across four datasets
and three base LLMs, including confidence
complementarity.
\end{itemize}

\section{Related Work}
\label{sec:related_work}

\paragraph{Stance Detection and Evidence-Based Reasoning.}
Stance detection identifies a text's position toward
a target, including implicit stances
\cite{mohammad-etal-2016-semeval}.
Prior work investigates target-aware modeling,
contextual knowledge, and cross-target transfer
\cite{garg-caragea-2024-stanceformer,
beck-etal-2023-robust,zhang-etal-2024-llm-driven,
zhang-etal-2023-cross-lingual,allaway2020zeroshot}.
Benchmarks span political and public-health discourse
\cite{li-etal-2021-p,glandt-etal-2021-stance},
while recent work explores open-target generalization,
multi-path reasoning, and retrieval-augmented verification
\cite{akash-etal-2025-large,zhang-etal-2025-mprf,
zhang-etal-2025-mpvstance}.
MPVStance \cite{zhang-etal-2025-mpvstance}
uses multi-perspective verification and retrieved
evidence to assess and revise stance responses.
Unlike MPVStance, B-CARE ranks errors in fixed
predictions without revising them.
Joint stance and rumor-veracity modeling has
also been studied \cite{yang-etal-2024-reinforcement}.
External evidence does not consistently improve
 prediction \cite{nguyen-kim-2025-external},
and zero-shot biases complicate reliability
assessment \cite{dubreuil-etal-2025-stereotypes}.

\paragraph{Uncertainty, Consistency, and Error Detection.}
Uncertainty quantification for LLMs encompasses
both white-box and black-box approaches,
including probability-based and consistency-based
methods \cite{shelmanov2025tutorial}.
Reliability assessment has been studied through
fact verification \cite{guan-etal-2024-language}, token-level and
 semantic consistency
\cite{wastl-etal-2025-uzh,
gupta-etal-2025-consistency},
and belief-state-based hallucination detection
\cite{zhang-etal-2025-bold}.
Sampling-based methods include SelfCheckGPT
\cite{manakul-etal-2023-selfcheckgpt}.
Chain-of-Verification uses independently answered
questions to verify generated content
\cite{dhuliawala-etal-2024-chain}.  B-CARE instead explicitly compares graded evidence for predicted and competing stance classes.

% Other approaches include self-evaluation
% \cite{kadavath2022language}, semantic entropy
% \cite{kuhn2023semantic,farquhar2024detecting},
% pairwise semantic similarity
% \cite{nguyen-etal-2025-beyond},
% and evidential uncertainty
% \cite{kunitomo-jacquin-etal-2026-evidential}.
% Semantic cross-checking and uncertainty-triggered
% verification provide additional reliability signals
% \cite{zhang-etal-2023-sac3,xue2025verify}.
% Cross-model agreement and disagreement have
% also been investigated
% \cite{goel-etal-2025-zero,hamidieh2026crossmodel},
% although consistent predictions can still be
% incorrect \cite{tan-etal-2025-consistent}.
% B-CARE instead explicitly compares graded
% evidence for predicted and competing stance classes.

\paragraph{Prediction-Blind Verification and Error Ranking.}
Confidence-based error detection and selective
classification provide established approaches
to prioritizing predictions for review
\cite{hendrycks2017baseline,geifman2017selective,
varshney2022selective}.
Independent verification and limitations of
self-consistency have also been investigated
\cite{dhuliawala-etal-2024-chain,
tan-etal-2025-consistent}.
MARCH \cite{li-etal-2026-march} employs a
solver--proposer--checker framework for
hallucination mitigation. Its checker verifies
claims derived from the solver's response against
retrieved evidence without seeing the original
answer. Thus, answer-blind verification is not
itself unique to B-CARE.

B-CARE obtains graded evidence for every stance
class directly from the target and text, without
observing the base prediction or deriving
verification questions from it. This evidence
is contrasted with each base prediction to rank
potential errors, rather than revise predictions
or verify response-derived factual claims.
We compare prediction-blind and prediction-aware
verification, graded and binary evidence, and
error prioritization under fixed review budgets.
The evidence profile is reusable across base
classifiers without training a separate error detector.

\section{Methodology}
\label{sec:method}
\begin{figure*}[t]
\centering
\includegraphics[width=0.98\textwidth]
{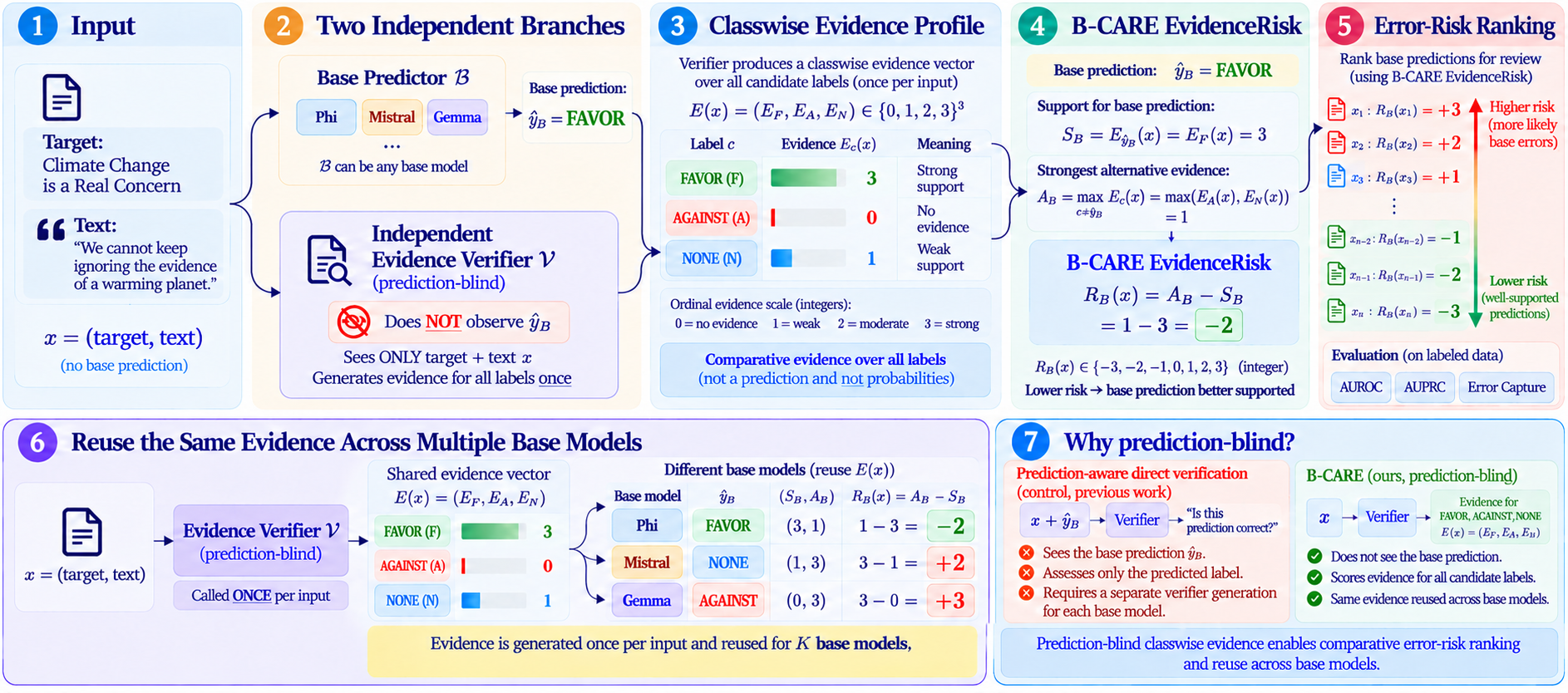}
\caption{
\textbf{Overview of B-CARE.}
An independent, prediction-blind LLM verifier assigns
ordinal evidence scores to all candidate stance labels
without observing the base prediction.
B-CARE contrasts evidence supporting the predicted
label with its strongest alternative to produce an
ordinal error-risk score for review prioritization.
The evidence profile can be reused across multiple
base classifiers without additional verifier generations.
}
\label{fig:bcare_framework}
\end{figure*}
Figure~\ref{fig:bcare_framework} illustrates the B-CARE
framework. A base classifier predicts the stance of
a target--text pair, while an independent verifier
assesses textual evidence for each candidate class
without observing the prediction. B-CARE then
compares the predicted class with its strongest
alternative to rank potentially incorrect predictions
for review.

\subsection{Error Ranking}
\label{sec:error_ranking}

Let $x=(T,X)$ denote an input comprising a target
$T$ and text $X$. A base classifier $B$ produces
a stance prediction
\begin{equation}
\hat{y}_B=f_B(x),\qquad
\hat{y}_B\in\mathcal{Y},
\label{eq:base_prediction}
\end{equation}
where $\mathcal{Y}$ denotes the dataset-specific
label set. SemEval-2016, VaccineStance, and
COVID-19-Stance use
$\mathcal{Y}=\{\textsc{favor},\textsc{against},
\textsc{none}\}$, while P-Stance uses
$\mathcal{Y}=\{\textsc{favor},\textsc{against}\}$.
The COVID-19-Stance label \textsc{neutral}
is mapped to \textsc{none}.

Given the gold label $y$, we define the
evaluation-only error indicator
\begin{equation}
z_B(x)=\mathbb{I}[\hat{y}_B\neq y].
\label{eq:error_indicator}
\end{equation}
Gold labels are unavailable to the risk estimator
at inference time. Our objective is to assign
each existing prediction a scalar risk score
$R_B(x)$, with larger values indicating higher
priority for review. Unlike stance classification,
this task does not generate or correct labels.
The risk score is not interpreted as a calibrated
probability of error.

\subsection{Prediction-Blind Classwise Evidence}
\label{sec:comparative_evidence}

B-CARE employs Llama-3.3-70B-Instruct through
OpenRouter as a fixed external evidence verifier
$V$, with temperature $0$ and a maximum of
80 output tokens per request. One independent
request produces a complete classwise evidence
profile. Frozen ternary and binary prompt variants
are used for the corresponding label sets,
with a common evidence scale and output schema.

Given the target, text, and candidate labels,
the verifier produces an ordinal evidence profile
\begin{equation}
E(x)=\bigl(E_c(x)\bigr)_{c\in\mathcal{Y}},
\label{eq:evidence_vector}
\end{equation}
where $E_c(x)\in\{0,1,2,3\}$.
The scale represents increasing textual support:
no evidence (0), weak or ambiguous evidence (1),
moderate evidence (2), and strong, direct
evidence (3).

The verifier evaluates textual evidence for
each candidate stance and is explicitly instructed
not to output a final stance prediction.
It receives neither the base prediction nor its
confidence, logits, or model identity. Gold labels
and correctness indicators are also withheld.
Thus, evidence generation depends on the input
and label set rather than on the base prediction.

For a base prediction $\hat{y}_B$, we define
the evidence supporting that prediction as
\begin{equation}
S_B(x)=E_{\hat{y}_B}(x).
\label{eq:predicted_support}
\end{equation}
The strongest competing evidence is
\begin{equation}
A_B(x)=
\max_{\substack{c\in\mathcal{Y}\\
c\neq\hat{y}_B}}E_c(x).
\label{eq:alternative_support}
\end{equation}
B-CARE then assigns the comparative
evidence risk
\begin{equation}
R_B(x)=A_B(x)-S_B(x).
\label{eq:risk}
\end{equation}

Since evidence scores lie in $\{0,1,2,3\}$,
the risk belongs to $\{-3,-2,\ldots,3\}$.
Positive scores indicate stronger evidence
for an alternative stance, whereas negative
scores indicate stronger support for the
predicted label. A zero score indicates equal
evidence for the predicted class and its
strongest alternative, not necessarily equal
support across all candidate classes.
Predictions are ranked in descending order
of $R_B(x)$.
For review budget $q$, we review
$k=\max(1,\mathrm{round}(qN))$ predictions per
dataset--model setting.

% Because the risk score takes only seven discrete
% values, tied predictions are common. AUROC uses
% midrank handling, while AUPRC uses average
% precision. For review budget $q$, we select
% $k=\max(1,\operatorname{round}(qN))$ predictions
% per dataset--model setting. When the budget
% boundary falls within a tied risk group,
% ErrorCapture uses expected capture under
% uniform selection within that group, avoiding
% arbitrary row-order tie-breaking.

% No learned error detector, dataset-specific
% calibration model, or optimized threshold
% is required. The verifier provides comparative
% evidence for ranking errors rather than replacing
% the original predictions.

\subsection{Prediction Blindness and Evidence Reuse}
\label{sec:blindness}

Prediction blindness separates evidence assessment
from base predictions and enables evidence reuse
across classifiers.

\paragraph{Prediction exposure.}
A prediction-aware verifier may condition its
assessment on the predicted label, potentially
introducing confirmation bias. B-CARE instead
generates evidence without observing that label.
We examine this distinction in
Section~\ref{sec:prediction_blindness}, comparing blind
and prediction-aware verification under the
same verifier backbone, evidence scale, output
schema, and risk formula. The prediction-aware
variant additionally receives the base prediction.

\paragraph{Evidence reuse.}
Because classwise evidence depends only on the
target--text input and label set, one verifier
generation can serve multiple base classifiers
evaluated on the same example. Their respective
risk scores are computed offline using the
shared evidence vector.

% Prediction-aware verification can also reuse
% cached responses when base classifiers predict
% the same label. Across 5{,}206 unique examples,
% B-CARE requires 5{,}206 logical verifier
% generations, compared with 6{,}544 for a cached
% prediction-aware alternative, a reduction of
% approximately 20.4\%. These counts exclude
% retries and reflect logical generations rather
% than total inference cost, token usage, or latency.

Prediction-aware verification can also cache
responses when base classifiers agree; across
5,206 examples, B-CARE requires 5,206 verifier
generations versus 6,544 for a cached
prediction-aware alternative (20.4\% fewer;
retries, tokens, and latency excluded).

\section{Experimental Setup}
\label{sec:experimental_setup}

\subsection{Datasets and Models}
\label{sec:datasets_models}

We evaluate B-CARE on four stance benchmarks
with 5,206 unique test examples across 13 targets:
SemEval-2016 Task 6A (five targets, excluding
Donald Trump) \cite{mohammad-etal-2016-semeval},
VaccineStance (1,000 examples)
\cite{sabbaghan2026moprosd},
COVID-19-Stance (four targets, 200 each)
\cite{glandt-etal-2021-stance}, and P-Stance
(2,157 examples, three targets)
\cite{li-etal-2021-p}. The first three use
\textsc{favor}, \textsc{against}, and
\textsc{none} (mapping neutral to
\textsc{none}); P-Stance is binary.
Target statistics appear in
Appendix~\ref{app:dataset_statistics}.

Phi-4 (14B) \cite{abdin2024phi4},
Mistral Small 24B Instruct (2501)
\cite{mistralai2025small2501}, and
Gemma 3 (27B) \cite{gemmateam2025gemma3}
were run on a university HPC cluster with revision-pinned
Hugging Face checkpoints, local Transformers
inference, and frozen stance prompts.

Across the 12 dataset--model settings,
B-CARE uses Llama-3.3-70B-Instruct
\cite{grattafiori2024llama3,meta2024llama33} via OpenRouter
as a fixed prediction-blind verifier
(temperature 0). A separate set of
2,000 validation examples (500 per dataset)
was used for method development and validation.
The verifier prompt, ordinal scale, output
schema, and inference settings were finalized
before test evaluation. Exact model revisions
and configurations are provided in the
reproducibility materials.
The verifier prompts and prediction-exposure
controls are detailed in
Appendix~\ref{app:verifier_prompts}.

No supervised error-risk training or
calibration is required. Training examples
are used only for specified baselines,
including 20 few-shot examples for
$p(\mathrm{True})$.

\subsection{Baselines and Controlled Variants}
\label{sec:baselines}

We compare against uncertainty,
self-evaluation, sampling, and verification
baselines and controlled variants.

\paragraph{Base-model uncertainty.}
Maximum Softmax Probability (MSP)
\cite{hendrycks2017baseline} uses one minus
the maximum class probability; Predictive
Entropy (PE) \cite{malinin2018predictive}
measures closed-set stance uncertainty.

\paragraph{Self-evaluation.}
$p(\mathrm{True})$
\cite{kadavath2022language,farquhar2024detecting}
estimates the correctness of the base prediction
following the Semantic Uncertainty implementation.
Our Direct Self-Verification (DSV) control uses a
separate True/False correctness prompt.

\paragraph{Sampling-based uncertainty.}
Label Entropy (LE) measures entropy over
sampled stance labels; Variation Ratio (VR)
measures disagreement with the modal label.
Semantic Entropy (SE)
\cite{kuhn2023semantic,farquhar2024detecting}
groups semantically equivalent responses;
Evidential Semantic Entropy (EVSE)
\cite{kunitomo-jacquin-etal-2026-evidential}
extends this through evidential modeling.

\paragraph{Comparative evidence controls.}
Same-Model Comparative Evidence applies
B-CARE's classwise evidence scoring and
$A_B-S_B$ formula using the base model
as verifier. Evidence-Label Disagreement
(LD) uses the external verifier's classwise
evidence to check whether the base prediction
uniquely attains the maximum evidence score;
ties are counted as disagreement.
Same-Model Label Disagreement applies the
same unique-maximum rule to evidence
generated by the base model.

\paragraph{Model-disagreement controls.}
Cross-Base Disagreement (CBD) measures
disagreement among the three base models;
Llama Label Disagreement compares the
base prediction with an independently
generated external Llama stance label.

\paragraph{External verification controls.}
External Direct Correctness Verification
(EDCV) exposes the base prediction for
direct correctness assessment.
Prediction-Aware Comparative Evidence
(PACE) retains B-CARE's verifier,
ordinal scale, output schema, and
$A_B-S_B$ formula but reveals the base
prediction, isolating prediction exposure.
Random ranking provides a chance reference;
standalone Llama stance classification
is evaluated separately.

\paragraph{Component and sensitivity analyses.}
We compare support-only ($-S_B$),
alternative-only ($A_B$), and comparative
($A_B-S_B$) scores, plus lexicographic
ranking, softmax-based support, and
normalized evidence share. A random-label
exposure control tests the effect of
revealing an unrelated candidate label.
These analyses are separate from primary
baseline comparisons.
Full implementation details for all baselines
and controlled variants are provided in
Appendix~\ref{app:baseline_implementation}.

\subsection{Metrics and Statistical Analysis}
\label{sec:metrics}

Our primary metric is error-ranking AUROC,
with incorrect predictions as positives.
We also report AUPRC, error prevalence,
area under the risk--coverage curve (AURC),
excess AURC (E-AURC), and ErrorCapture@$q$
(EC@$q$), the fraction of existing errors
captured by reviewing the highest-risk
$q$ fraction of predictions.

Because $R_B(x)\in\{-3,-2,\ldots,3\}$,
ties are common. AUROC uses midranks,
AUPRC uses average precision over score
thresholds, and EC@$q$ uses expected
fractional allocation at budget-boundary
ties to avoid arbitrary ordering.

Paired within-setting bootstrap comparisons
use 10,000 matched-example resamples with
predefined seeds. Means and standard
deviations across the 12 settings reflect
between-setting variation, not run-to-run
uncertainty. Win counts are descriptive
because settings are not independent
experimental replications. 
Tie handling for AURC and E-AURC is described in Appendix \ref{app:baseline_implementation}.

% \paragraph{Tie Handling in AURC and E-AURC.}
% For risk--coverage evaluation, instances are ranked
% in ascending order of estimated error risk. When
% multiple instances receive identical risk scores,
% ties are handled using expected fractional allocation.
% Specifically, for a group of $g$ tied instances
% containing $e$ errors, each successive coverage step
% receives an expected error contribution of $e/g$.
% AURC is computed by averaging the selective error
% rate across all coverage levels. E-AURC is obtained
% by subtracting the oracle AURC, which ranks all
% correct predictions before incorrect predictions.
% This treatment avoids dependence on arbitrary
% ordering within tied groups.

% Required packages in the preamble:
% \usepackage{graphicx}
% \usepackage{booktabs}
% \usepackage{amsmath}

\section{Main Results}
\label{sec:main_results}

\subsection{Overall Error-Ranking Performance}
\label{sec:overall_performance}
\begin{figure*}[!t]
\centering
\hspace{-15mm}
\includegraphics[width=1.05\textwidth]
{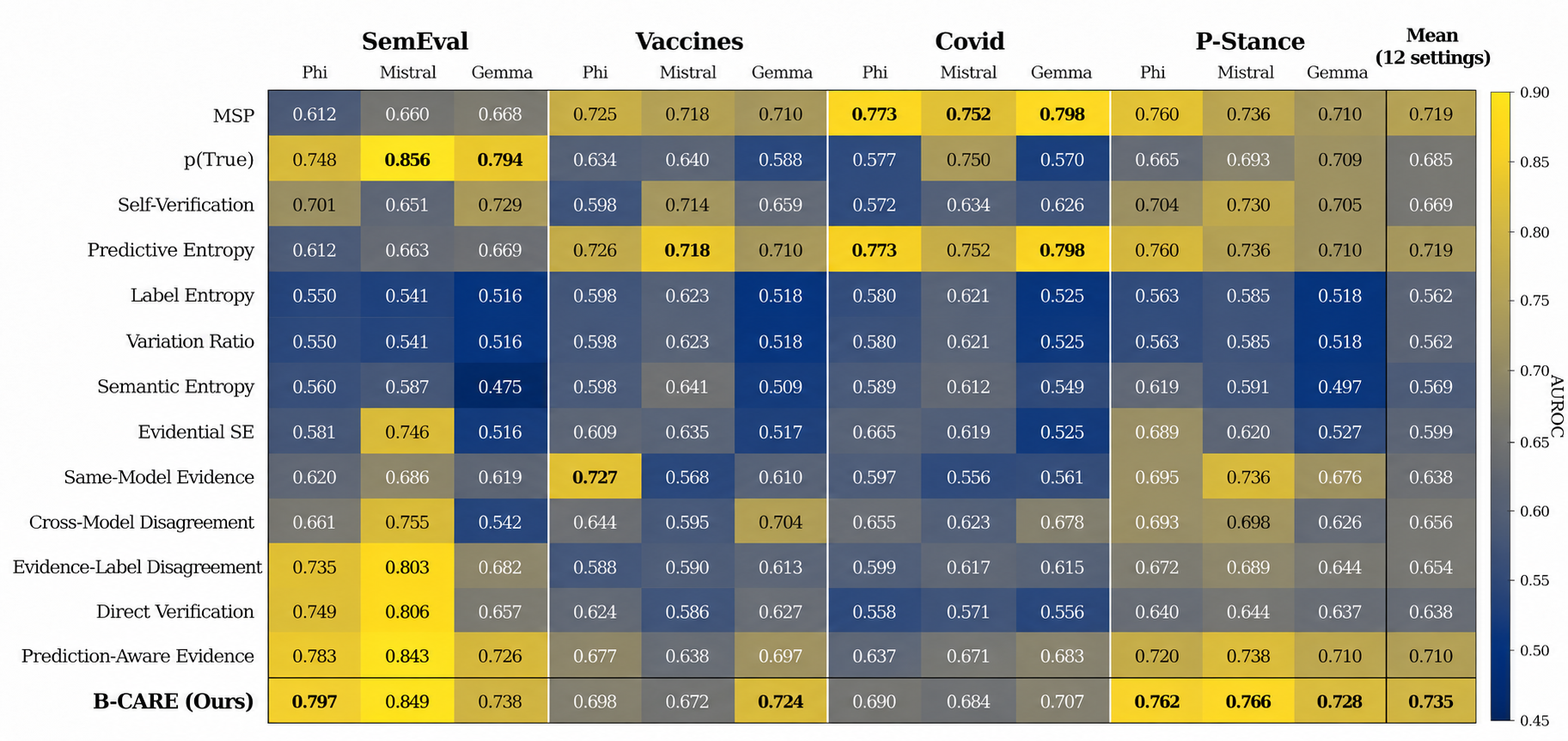}

\caption{
\textbf{AUROC comparison across 12 dataset--model settings.}
B-CARE and 13 competing error-ranking methods
are evaluated on four stance-detection benchmarks
and three base models, treating incorrect predictions
as positive. Cell values are shown to three decimal
places; boldface identifies the highest unrounded
AUROC within each setting. The rightmost column
reports the mean across all 12 settings.
B-CARE achieves the highest mean AUROC (0.735)
among the 14 methods in this primary comparison
and leads in five settings.
}
\label{fig:auroc_heatmap}
\end{figure*}
Figure~\ref{fig:auroc_heatmap} compares B-CARE
with 13 competing error-ranking methods across
12 dataset--model settings, covering four
stance-detection benchmarks and three base models.
AUROC measures how effectively each method ranks
incorrect predictions above correct ones,
independently of a review budget or decision threshold.

B-CARE achieves the highest AUROC in five of
the 12 settings: SemEval--Phi, Vaccines--Gemma,
and all three P-Stance settings. Notable AUROC
values include 0.797 on SemEval--Phi, 0.849 on
SemEval--Mistral, and 0.766 on P-Stance--Mistral.
Its performance across all three P-Stance
base models suggests that comparative classwise
evidence can provide a useful error-ranking
signal across different model architectures.

Performance nevertheless varies substantially
across datasets and models. On SemEval, B-CARE
performs strongly against most alternatives,
although $p(\mathrm{True})$ achieves higher AUROC
for Mistral and Gemma. On Vaccines, B-CARE
leads with Gemma, while Same-Model Evidence
achieves the highest AUROC for Phi and
confidence-based methods remain competitive
for Mistral. On COVID, MSP and Predictive
Entropy outperform B-CARE across all three
base models. Thus, the benefits of comparative
evidence are setting-dependent rather than
universal.

B-CARE also outperforms Direct Verification
(EDCV) and Prediction-Aware Comparative Evidence
(PACE) in AUROC across the evaluated settings.
Unlike direct correctness judgments or evidence
conditioned on the predicted label, B-CARE
compares prediction-independent evidence
for competing stance classes.

A separate matched evaluation against verifier
probabilities is reported in
Section~\ref{sec:verifier_probabilities}.
Additional label-disagreement, random-ranking,
and normalized evidence-share controls are
summarized in
Appendix~\ref{app:additional_controls}
(Table~\ref{tab:additional_controls}).

\subsection{Complementary Evaluation Metrics}
\label{sec:complementary_metrics}

\begin{table*}[!t]
\centering
\begingroup
\small
\setlength{\tabcolsep}{5pt}
\renewcommand{\arraystretch}{1.08}

\resizebox{\textwidth}{!}{%
\begin{tabular}{lccccc}
\toprule
\textbf{Method} &
\textbf{AUPRC $\uparrow$} &
\textbf{E-AURC $\downarrow$} &
\textbf{EC@10\% $\uparrow$} &
\textbf{EC@20\% $\uparrow$} &
\textbf{EC@30\% $\uparrow$} \\
\midrule

MSP
& $0.500 \pm 0.093$
& $0.121 \pm 0.049$
& $20.33 \pm 4.67$
& $37.48 \pm 7.10$
& $51.26 \pm 7.66$ \\

$p(\mathrm{True})$
& $0.468 \pm 0.139$
& $0.132 \pm 0.056$
& $18.53 \pm 5.98$
& $32.62 \pm 9.24$
& $46.16 \pm 9.53$ \\

Self-Verification
& $0.429 \pm 0.069$
& $0.138 \pm 0.050$
& $16.62 \pm 4.97$
& $31.50 \pm 6.42$
& $44.41 \pm 7.22$ \\

Predictive Entropy
& $0.503 \pm 0.092$
& $\mathbf{0.119} \pm 0.050$
& $20.20 \pm 4.54$
& $37.45 \pm 7.08$
& $51.25 \pm 7.58$ \\

Label Entropy
& $0.353 \pm 0.096$
& $0.218 \pm 0.058$
& $16.78 \pm 4.21$
& $27.92 \pm 5.51$
& $36.94 \pm 4.82$ \\

Variation Ratio
& $0.353 \pm 0.096$
& $0.218 \pm 0.058$
& $16.78 \pm 4.21$
& $27.92 \pm 5.51$
& $36.94 \pm 4.82$ \\

Semantic Entropy
& $0.379 \pm 0.099$
& $0.218 \pm 0.060$
& $17.20 \pm 5.56$
& $28.23 \pm 7.16$
& $37.90 \pm 7.10$ \\

Evidential SE
& $0.405 \pm 0.110$
& $0.191 \pm 0.067$
& $17.77 \pm 4.80$
& $29.61 \pm 5.60$
& $39.06 \pm 5.79$ \\

Same-Model Evidence
& $0.407 \pm 0.098$
& $0.178 \pm 0.060$
& $17.65 \pm 7.57$
& $32.91 \pm 9.84$
& $45.81 \pm 9.97$ \\

Cross-Model Disagreement
& $0.441 \pm 0.121$
& $0.172 \pm 0.043$
& $21.37 \pm 6.63$
& $36.56 \pm 7.70$
& $49.12 \pm 6.02$ \\

Evidence-Label Disagreement
& $0.441 \pm 0.124$
& $0.174 \pm 0.052$
& $22.61 \pm 8.29$
& $38.47 \pm 6.61$
& $48.29 \pm 6.83$ \\

Direct Verification
& $0.395 \pm 0.128$
& $0.170 \pm 0.055$
& $15.06 \pm 6.51$
& $31.16 \pm 6.73$
& $44.56 \pm 6.12$ \\

Prediction-Aware Evidence
& $0.508 \pm 0.127$
& $0.142 \pm 0.042$
& $23.32 \pm 7.14$
& $39.11 \pm 7.84$
& $53.64 \pm 7.64$ \\

\midrule
\textbf{B-CARE (Ours)}
& $\mathbf{0.524} \pm 0.121$
& $0.125 \pm 0.035$
& $\mathbf{23.85} \pm 8.11$
& $\mathbf{41.06} \pm 8.96$
& $\mathbf{54.35} \pm 7.52$ \\
\bottomrule
\end{tabular}%
}
\endgroup

\caption{
Complementary error-ranking performance
(mean $\pm$ sample standard deviation across
12 dataset--model settings).
Higher AUPRC and EC and lower E-AURC are better.
EC is expressed as a percentage and uses expected
fractional allocation at budget-boundary ties.
Bold indicates the best mean in each column.
}
\label{tab:complementary_results}
\end{table*}
Beyond AUROC, we evaluate AUPRC, excess area
under the risk--coverage curve (E-AURC), and
ErrorCapture (EC) at review budgets of 10\%,
20\%, and 30\%. Table~\ref{tab:complementary_results}
reports mean performance across all 12
dataset--model settings.
B-CARE achieves the highest mean AUPRC (0.524)
and EC at all three budgets (23.85\%, 41.06\%,
and 54.35\%, respectively), indicating effective
error prioritization under limited review budgets.
However, Predictive Entropy achieves a lower
E-AURC (0.119 versus 0.125). These findings
demonstrate advantages in specific review
metrics without implying uniform superiority.

\subsection{Statistical Significance}
\label{sec:statistical_significance}

We compare AUROC using paired, example-level
bootstrap resampling with unadjusted percentile
95\% confidence intervals across 156 comparisons.
B-CARE shows positive AUROC differences with
intervals excluding zero in all 12 settings
against Direct Verification, Evidential SE,
Label Entropy, Variation Ratio, and
Evidence-Label Disagreement, and in 11/12
settings against Same-Model Evidence and
Semantic Entropy. The only nonsignificant
comparisons for these two baselines occur
on Vaccines with Phi and Vaccines with Mistral,
respectively.
Against Prediction-Aware Evidence, B-CARE
records seven significant wins, no significant
losses, and five nonsignificant differences.
Comparisons with MSP and Predictive Entropy
each yield four wins, four losses, and four
nonsignificant differences.

The Same-Model Evidence comparison preserves
the evidence rubric and ranking rule but
changes the verifier and inference environment;
the observed performance gap therefore cannot
be attributed solely to verifier capacity.
These findings are based on unadjusted intervals
and do not establish family-wise significance
across all comparisons. Complete results appear
in Appendix~\ref{app:statistical_significance}.

\subsection{Complementarity with Predictive Entropy}
\label{sec:complementarity}

We examine complementarity between B-CARE
and Predictive Entropy (PE). Across 12
dataset--model settings, the sets of actual
errors captured within B-CARE's and PE's
independently selected top-20\% review sets
have a mean Jaccard overlap of 0.406, while
their union captures 55.5\% of all errors (Table~\ref{tab:complementarity_overlap}; see also Figure~\ref{fig:complementarity} in Appendix~\ref{app:complementarity}).
Importantly, this union exceeds a single
20\% review budget.

% \begin{figure}[t]
% \centering
% \includegraphics[width=\columnwidth]%
% {figures/BCARE_complementarity_crisp_labels (1).png}
% \caption{
% Complementarity between B-CARE and Predictive
% Entropy (PE) across 12 dataset--model settings.
% \textbf{Top:} AUROC of B-CARE, PE, and the
% exploratory equal-weight hybrid.
% \textbf{Bottom:} Errors captured exclusively
% by B-CARE, exclusively by PE, or by both
% within separate top-20\% review sets.
% Segment heights represent percentages of
% all errors; labels indicate error counts.
% }
% \label{fig:complementarity}
% \end{figure}

An exploratory, untuned 0.5/0.5 average of
within-setting normalized ranks achieves
a mean AUROC of 0.760, compared with 0.735
for B-CARE and 0.719 for PE. The hybrid
outperforms B-CARE in 9/12 settings but
underperforms in all three SemEval settings.
These findings suggest complementary,
dataset-dependent signals rather than
uniform improvements. The hybrid is not
part of the frozen B-CARE method.
Additional results appear in
Appendix~\ref{app:complementarity}.

\section{Ablation Studies}
\label{sec:ablation}

\subsection{Component Ablation of the B-CARE Score}
\label{sec:score_ablation}

\begin{figure}[H]
    \centering
    \includegraphics[width=\columnwidth]
    {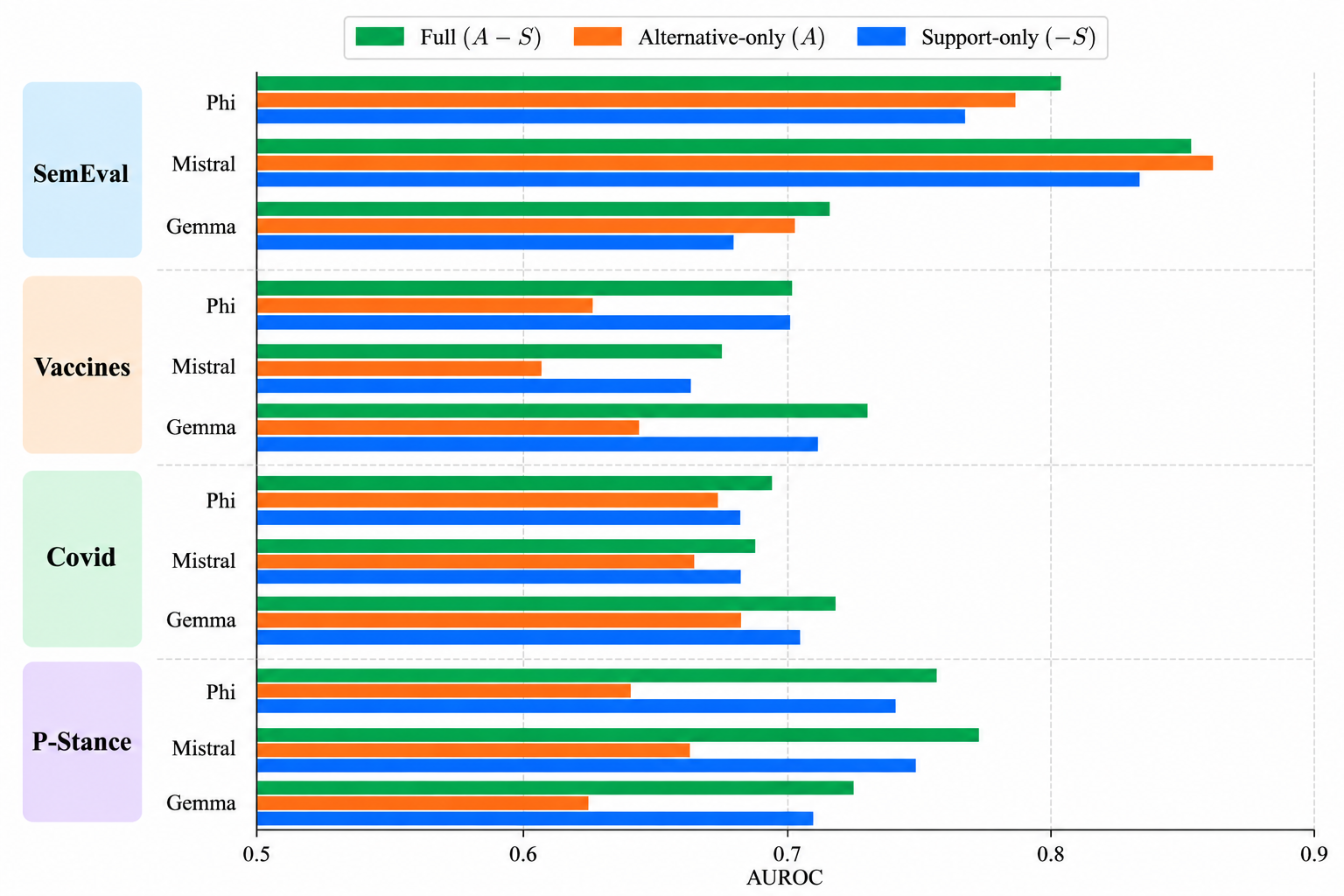}
    \caption{Component ablation of the B-CARE
    score across 12 dataset--model settings.
    Higher AUROC indicates better error ranking.}
    \label{fig:score_ablation}
\end{figure}

We examine the contribution of each component
of the B-CARE score by comparing the full
formulation ($A-S$) with support-only ($-S$)
and alternative-only ($A$) variants.
As shown in Figure~\ref{fig:score_ablation},
the full score achieves the highest mean AUROC
(0.735), compared with 0.728 for $-S$ and
0.686 for $A$. It outperforms the two variants
in 11/12 and 10/12 settings, respectively.
Paired bootstrap comparisons identify
significant gains in 6/12 and 9/12 settings,
with no significant losses. These results
indicate that predicted-label support provides
a strong ranking signal, while competing-class
evidence offers additional benefits.
Detailed results are provided in
Appendix~\ref{app:score_ablation}.

\subsection{Ordinal versus Continuous Evidence Scoring}
\label{sec:continuous_ablation}

\begin{figure}[t]
    \centering
    \includegraphics[width=\columnwidth]
    {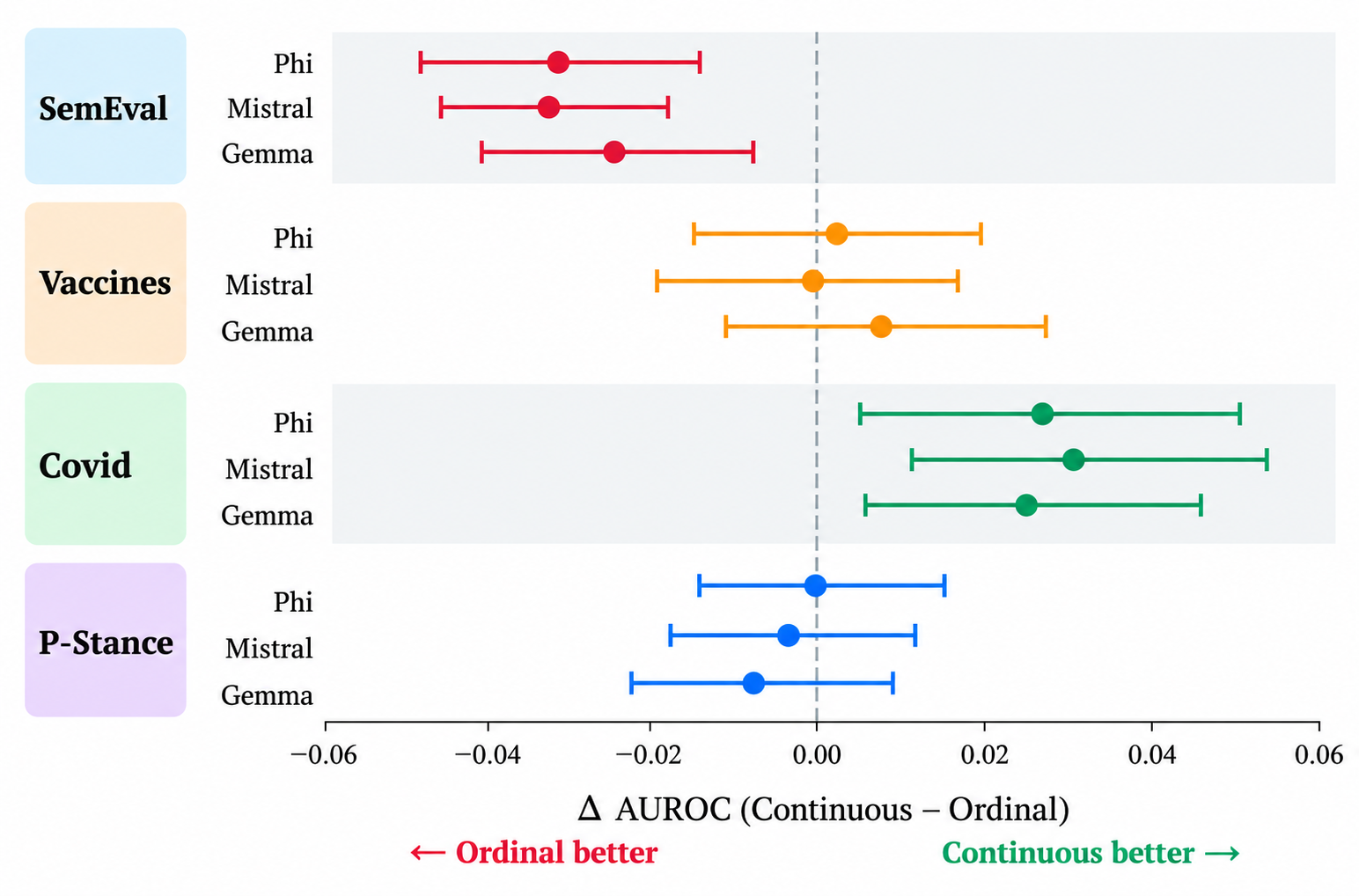}
    \caption{
Ordinal versus continuous evidence scoring across
12 dataset--model settings.
Dots show AUROC differences (continuous minus ordinal);
horizontal bars indicate unadjusted 95\% paired bootstrap
confidence intervals (10,000 resamples).
Positive values favor continuous scoring, negative values
favor ordinal scoring, and intervals excluding zero
indicate differences at the unadjusted 95\% level.
}
    \label{fig:continuous_ordinal}
\end{figure}

We test whether continuous evidence scores improve
on ordinal grades. The same fixed, prediction-blind
Llama verifier is prompted for real-valued classwise
scores on $[0,3]$, with the same
$R_B = A_B - S_B$ formulation. The two variants
achieve nearly identical mean AUROC (0.7346 ordinal
vs.\ 0.7345 continuous). As shown in
Figure~\ref{fig:continuous_ordinal}, ordinal
scoring is better on SemEval for all three base
models, continuous scoring is better on COVID
(intervals excluding zero in two of three settings),
and neither has a clear advantage on Vaccines or
P-Stance. Finer granularity therefore yields no
consistent gain, supporting ordinal scoring as a
competitive and interpretable choice.

% We investigate whether replacing B-CARE's ordinal evidence
% grades with continuous evidence scores improves error ranking.
% In the continuous variant, the same fixed, prediction-blind
% Llama verifier is prompted to directly generate real-valued
% classwise evidence scores on $[0,3]$, instead of the
% ordinal grades $\{0,1,2,3\}$.
% Both variants use the same relative-evidence formulation,
% $R_B=A_B-S_B$, while holding the verifier backbone,
% prediction blindness, and relative-evidence formulation
% fixed. This comparison evaluates whether continuous
% evidence elicitation provides an advantage over the
% ordinal formulation.

% Across 12 dataset--model settings, the two variants achieve
% nearly identical mean AUROC (0.7346 for ordinal versus
% 0.7345 for continuous scoring).
% As shown in Figure~\ref{fig:continuous_ordinal},
% ordinal scoring performs better on SemEval across all three
% base models, whereas continuous scoring improves AUROC
% on Covid, with confidence intervals excluding zero in
% two of the three settings.
% Neither variant exhibits a clear advantage on Vaccines
% or P-Stance.
% Although continuous scores provide finer-grained risk
% rankings, this additional granularity does not yield a
% consistent improvement in error detection.
% These findings support ordinal evidence scoring as a
% competitive and interpretable design choice.

\subsection{Effect of Prediction Blindness}
\label{sec:prediction_blindness}

\begin{figure}[t]
    \centering
    \includegraphics[width=\columnwidth]
    {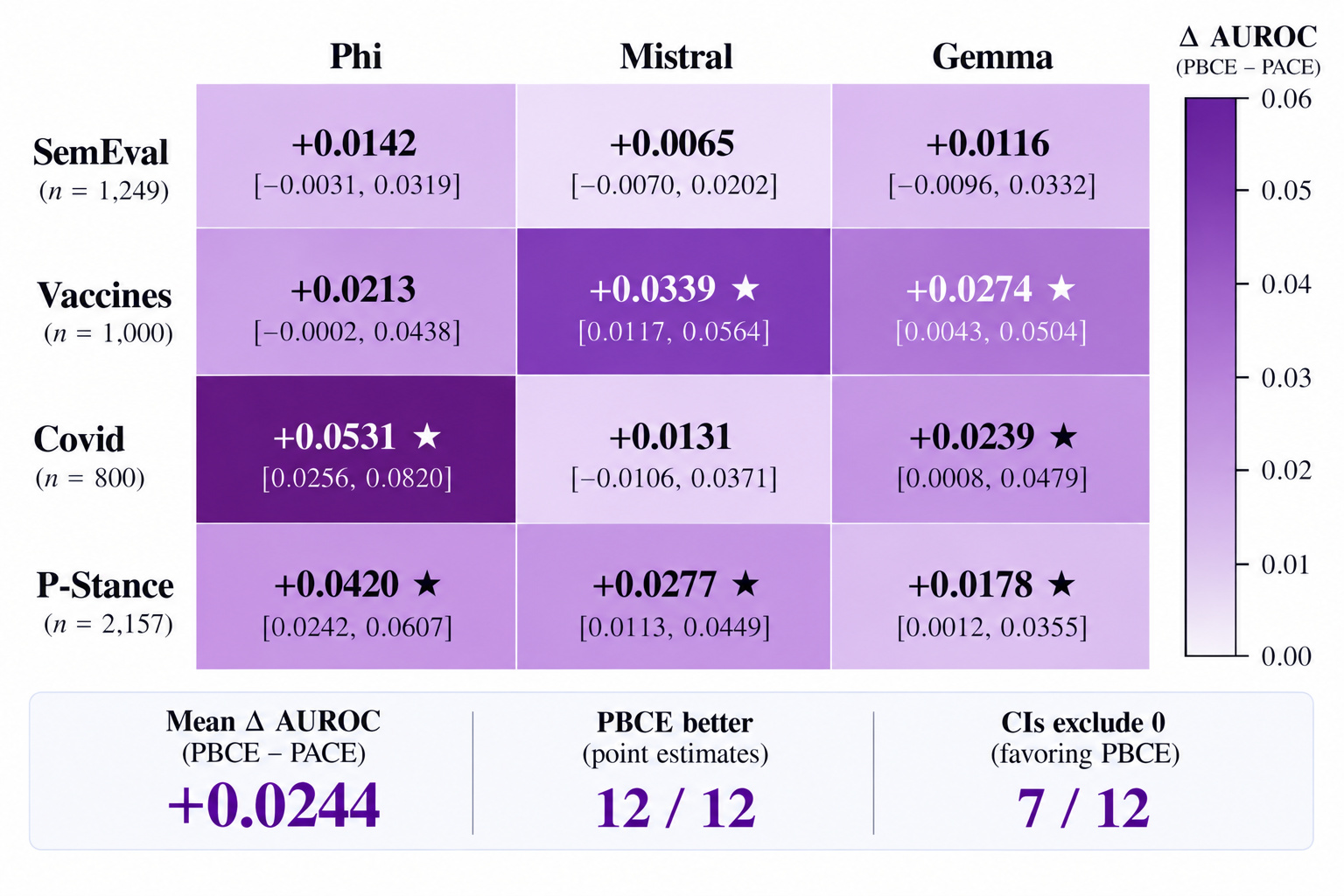}
    \caption{
    Effect of prediction blindness on error-ranking AUROC.
    Cells report the AUROC difference (PBCE minus PACE)
    with unadjusted 95\% paired-bootstrap confidence
    intervals (10,000 resamples).
    Darker purple indicates larger gains; stars denote
    intervals excluding zero.
    }
    \label{fig:prediction_blindness}
\end{figure}

We assess the role of prediction blindness by comparing
B-CARE's prediction-blind evidence (PBCE) with a
prediction-aware variant (PACE). Both use the same
Llama-3.3-70B verifier, evidence scale, and $A_B-S_B$
score; PACE additionally receives the base prediction.
As shown in Figure~\ref{fig:prediction_blindness},
PBCE achieves higher AUROC in all 12 dataset--model
settings, with a mean improvement of 0.0244.
Seven paired-bootstrap 95\% confidence intervals
exclude zero in favor of PBCE, with none favoring PACE.
These results support prediction blindness as a
beneficial design choice, although not every individual
improvement is statistically significant.

\subsection{Graded Evidence versus Evidence-Label Disagreement}
\label{sec:graded_binary_ablation}

% \begin{figure}[t]
% \centering
% \includegraphics[width=\columnwidth]
% {figures/AUROC-Gains-Heatmap-Summary.png}
% \caption{
% AUROC gains of graded B-CARE over binary
% Evidence-Label Disagreement across 12
% dataset--model settings. Cells show
% $\Delta$AUROC and unadjusted 95\%
% paired-bootstrap confidence intervals;
% stars denote intervals excluding zero.
% }
% \label{fig:graded_binary_ablation}
% \end{figure}

We assess whether graded relative evidence improves
error ranking beyond Evidence-Label Disagreement,
using identical frozen verifier outputs. As shown in Table \ref{tab:binary_full_auroc} (Appendix \ref{app:graded_binary_ablation}),
%Figure~\ref{fig:graded_binary_ablation},
B-CARE achieves higher AUROC in all 12 settings
(mean $\Delta$AUROC $= +0.0803$), with all 12
unadjusted 95\% paired-bootstrap confidence
intervals excluding zero.

These results show that preserving graded
relative evidence provides a stronger
error-ranking signal than binary
Evidence-Label Disagreement under the
evaluated settings. Detailed results and
additional analyses are provided in
Appendix~\ref{app:graded_binary_ablation}.

\subsection{Comparison with Verifier Probabilities}
\label{sec:verifier_probabilities}

% We additionally compare B-CARE with LlamaProb
% (LP), which ranks errors using
% $1-p_V(\hat{y}_B\mid x,t)$.
% Both methods were evaluated using the same
% self-hosted Llama-3.3-70B-Instruct verifier
% on RunPod, with identical test instances
% and frozen base predictions. This matched
% evaluation is separate from the primary
% 14-method comparison, which used a different
% inference environment.

% B-CARE achieves higher AUROC in 9 of the
% 12 settings, although the two methods
% have nearly identical mean AUROC
% (0.760 vs.\ 0.761).
% B-CARE also achieves higher mean AUPRC
% (0.603 vs.\ 0.592) and error capture
% at a 20\% review budget
% (0.444 vs.\ 0.428), whereas LlamaProb
% achieves lower mean E-AURC
% (0.108 vs.\ 0.116).

% Performance varies across datasets:
% B-CARE leads on VaccineStance and P-Stance,
% whereas LlamaProb leads on COVID-19-Stance.
% These results qualify the primary comparison:
% B-CARE achieves the highest mean AUROC
% among the 14 methods in the original
% evaluation, but verifier-probability ranking
% performs comparably on average in the
% separate matched experiment.

% Complete per-setting results,
% inference-environment provenance,
% and diagnostic analyses of COVID-19-Stance
% are provided in
% Appendix~\ref{app:runpod_matched}.

We also compare B-CARE with LlamaProb (LP), which
ranks errors by $1-p_V(\hat{y}_B\mid x,t)$ from the
same verifier. Both were evaluated on a self-hosted
Llama-3.3-70B-Instruct with identical test instances
and frozen base predictions; this matched evaluation
is separate from the 14-method comparison, which used
a different inference environment.
% B-CARE achieves higher AUROC in 9 of 12 settings,
% with nearly identical mean AUROC (0.760 vs.\ 0.761),
% higher mean AUPRC (0.603 vs.\ 0.592) and
% EC@20\% (0.444 vs.\ 0.428), and higher E-AURC
% (0.116 vs.\ 0.108). 
B-CARE achieves higher AUROC in 9 of 12 settings, with nearly identical mean AUROC (0.760 vs. 0.761). B-CARE achieves higher mean AUPRC (0.603 vs. 0.592) and EC@20\% (0.444 vs. 0.428), whereas LP achieves lower mean E-AURC (0.108 vs. 0.116).

B-CARE leads on VaccineStance
and P-Stance; LP leads on COVID-19-Stance.
B-CARE thus has the highest mean AUROC among the
14 methods in the primary evaluation, but
verifier-probability ranking performs comparably
on average. Per-setting results, inference-environment
provenance, and COVID-19-Stance diagnostics are in
Appendix~\ref{app:runpod_matched}.

\section{Conclusion}
\label{sec:conclusion}

We introduced B-CARE, a training-free framework
for ranking errors in frozen stance predictions
using prediction-blind, classwise ordinal
evidence from an external LLM verifier.
Across four datasets, three base models,
and 12 evaluation settings, B-CARE achieved
mean AUROC of 0.735 and AUPRC of 0.524,
capturing 41.06\% of existing errors within
a 20\% review budget. Controlled ablations
demonstrated the benefits of prediction
blindness and graded comparative evidence
over prediction-aware and binary-disagreement
alternatives. However, performance varied
across datasets, and B-CARE did not
consistently outperform strong confidence
baselines. These findings position B-CARE
as a complementary, reusable signal for
prioritizing human review rather than a
universal replacement for model uncertainty.
Future work will investigate adaptive
combinations with confidence-based signals,
finer-grained evidence scoring, and
generalization beyond stance detection.

\clearpage

\section*{Limitations}

B-CARE relies on the quality of classwise
evidence generated by an external LLM
verifier. Although prediction blindness
prevents direct exposure to the base
prediction, it does not eliminate shared
model biases or guarantee correct evidence.
The seven-level ordinal risk score also
produces ties that can limit ranking
resolution, particularly within large
groups of equally scored examples.

Our evaluation spans four English-language
stance datasets, 13 targets, and three
base LLMs, but uses a single external
verifier family. Generalization to other
languages, domains, tasks, and verifiers
remains untested. Performance varies across
datasets and does not consistently surpass
strong confidence-based baselines, suggesting
that comparative evidence and model
uncertainty capture complementary signals.

External verification incurs additional
inference cost relative to uncertainty
scores derived directly from base-model
outputs. Furthermore, the original
OpenRouter-hosted verifier lacked a pinned
backend checkpoint revision, limiting exact
output reproducibility. Finally, B-CARE
prioritizes potential errors rather than
correcting them; improvements from actual
human review have not been directly
evaluated.

\bibliography{references}

@misc{abdin2024phi4,
  title={Phi-4 Technical Report},
  author={Abdin, Marah and others},
  year={2024},
  eprint={2412.08905},
  archivePrefix={arXiv},
  url={https://arxiv.org/abs/2412.08905}
}

@inproceedings{akash-etal-2025-large,
  title={Can Large Language Models Address Open-Target Stance Detection?},
  author={Akash, Abu Ubaida and Fahmy, Ahmed and Trabelsi, Amine},
  booktitle={Findings of ACL 2025},
  year={2025},
  pages={971--985},
  publisher={Association for Computational Linguistics},
  doi={10.18653/v1/2025.findings-acl.54},
  url={https://aclanthology.org/2025.findings-acl.54/}
}

@inproceedings{allaway2020zeroshot,
  title={Zero-Shot Stance Detection: A Dataset and Model using Generalized Topic Representations},
  author={Allaway, Emily and McKeown, Kathleen},
  booktitle={Proceedings of the 2020 Conference on Empirical Methods in Natural Language Processing (EMNLP)},
  pages={8913--8931},
  year={2020},
  publisher={Association for Computational Linguistics},
  doi={10.18653/v1/2020.emnlp-main.717},
  url={https://aclanthology.org/2020.emnlp-main.717/}
}

@inproceedings{beck-etal-2023-robust,
  title={Robust Integration of Contextual Information for Cross-Target Stance Detection},
  author={Beck, Tilman and Waldis, Andreas and Gurevych, Iryna},
  booktitle={Proceedings of the 12th Joint Conference on Lexical and Computational Semantics (*SEM 2023)},
  pages={494--511},
  year={2023},
  publisher={Association for Computational Linguistics},
  doi={10.18653/v1/2023.starsem-1.43},
  url={https://aclanthology.org/2023.starsem-1.43/}
}

@inproceedings{dhuliawala-etal-2024-chain,
  title={Chain-of-Verification Reduces Hallucination in Large Language Models},
  author={Dhuliawala, Shehzaad and Komeili, Mojtaba and Xu, Jing and Raileanu, Roberta and Li, Xian and Celikyilmaz, Asli and Weston, Jason},
  booktitle={Findings of ACL 2024},
  year={2024},
  pages={3563--3578},
  doi={10.18653/v1/2024.findings-acl.212},
  url={https://aclanthology.org/2024.findings-acl.212/}
}

@inproceedings{dubreuil-etal-2025-stereotypes,
  title={Are Stereotypes Leading {LLM}s' Zero-Shot Stance Detection?},
  author={Dubreuil, Anthony and Gourru, Antoine and Largeron, Christine and Trabelsi, Amine},
  booktitle={Proceedings of the 2025 Conference on Empirical Methods in Natural Language Processing},
  pages={31517--31530},
  year={2025},
  publisher={Association for Computational Linguistics},
  doi={10.18653/v1/2025.emnlp-main.1605},
  url={https://aclanthology.org/2025.emnlp-main.1605/}
}

@article{farquhar2024detecting,
  title={Detecting hallucinations in large language models using semantic entropy},
  author={Farquhar, Sebastian and Kossen, Jannik and Kuhn, Lorenz and Gal, Yarin},
  journal={Nature},
  volume={630},
  pages={625--630},
  year={2024},
  doi={10.1038/s41586-024-07421-0},
  url={https://doi.org/10.1038/s41586-024-07421-0}
}

@inproceedings{garg-caragea-2024-stanceformer,
  title={Stanceformer: Target-Aware Transformer for Stance Detection},
  author={Garg, Krishna and Caragea, Cornelia},
  booktitle={Findings of the Association for Computational Linguistics: EMNLP 2024},
  pages={4969--4984},
  year={2024},
  publisher={Association for Computational Linguistics},
  doi={10.18653/v1/2024.findings-emnlp.286},
  url={https://aclanthology.org/2024.findings-emnlp.286/}
}

@misc{gemmateam2025gemma3,
  title={Gemma 3 Technical Report},
  author={{Gemma Team}},
  year={2025},
  eprint={2503.19786},
  archivePrefix={arXiv},
  url={https://arxiv.org/abs/2503.19786}
}

@inproceedings{glandt-etal-2021-stance,
  title={Stance Detection in {COVID-19} Tweets},
  author={Glandt, Kyle and Khanal, Sarthak and Li, Yingjie and Caragea, Doina and Caragea, Cornelia},
  booktitle={Proceedings of ACL-IJCNLP 2021},
  year={2021},
  pages={1596--1611},
  publisher={Association for Computational Linguistics},
  doi={10.18653/v1/2021.acl-long.127},
  url={https://aclanthology.org/2021.acl-long.127/}
}

@inproceedings{goel-etal-2025-zero,
  title={Zero-knowledge {LLM} hallucination detection and mitigation through fine-grained cross-model consistency},
  author={Goel, Aman and Schwartz, Daniel and Qi, Yanjun},
  booktitle={EMNLP 2025 Industry Track},
  year={2025},
  pages={1982--1999},
  doi={10.18653/v1/2025.emnlp-industry.139},
  url={https://aclanthology.org/2025.emnlp-industry.139/}
}

@article{grattafiori2024llama3,
  title={The Llama 3 Herd of Models},
  author={Grattafiori, Aaron and others},
  journal={arXiv preprint arXiv:2407.21783},
  year={2024},
  url={https://arxiv.org/abs/2407.21783}
}

@inproceedings{guan-etal-2024-language,
  title={Language Models Hallucinate, but May Excel at Fact Verification},
  author={Guan, Jian and Dodge, Jesse and Wadden, David and Huang, Minlie and Peng, Hao},
  booktitle={Proceedings of the 2024 Conference of the North American Chapter of the Association for Computational Linguistics: Human Language Technologies (Volume 1: Long Papers)},
  year={2024},
  doi={10.18653/v1/2024.naacl-long.62},
  url={https://aclanthology.org/2024.naacl-long.62/},
  pages={1090--1111},
  address={Mexico City, Mexico},
  publisher={Association for Computational Linguistics},
}

@inproceedings{gupta-etal-2025-consistency,
  title={Consistency Is the Key: Detecting Hallucinations in {LLM} Generated Text By Checking Inconsistencies About Key Facts},
  author={Gupta, Raavi and Panicker, Pranav Hari and Bhatia, Sumit and Ramakrishnan, Ganesh},
  booktitle={Proceedings of the 14th International Joint Conference on Natural Language Processing and the 4th Conference of the Asia-Pacific Chapter of the Association for Computational Linguistics},
  year={2025},
  doi={10.18653/v1/2025.findings-ijcnlp.129},
  url={https://aclanthology.org/2025.findings-ijcnlp.129/},
  pages={2053--2068},
  address={Mumbai, India},
  publisher={The Asian Federation of Natural Language Processing and The Association for Computational Linguistics},
}

@article{hamidieh2026crossmodel,
  title={Complementing Self-Consistency with Cross-Model Disagreement for Uncertainty Quantification},
  author={Hamidieh, Kimia and Thost, Veronika and Gerych, Walter and Yurochkin, Mikhail and Ghassemi, Marzyeh},
  journal={arXiv preprint arXiv:2604.17112},
  year={2026},
  url={https://arxiv.org/abs/2604.17112}
}

@inproceedings{hendrycks2017baseline,
  title={A Baseline for Detecting Misclassified and Out-of-Distribution Examples in Neural Networks},
  author={Hendrycks, Dan and Gimpel, Kevin},
  booktitle={International Conference on Learning Representations},
  year={2017},
  url={https://openreview.net/forum?id=Hkg4TI9xl}
}

@misc{kadavath2022language,
  title={Language Models (Mostly) Know What They Know},
  author={Kadavath, Saurav and Conerly, Tom and Askell, Amanda and Henighan, Tom and Drain, Dawn and Perez, Ethan and others},
  year={2022},
  eprint={2207.05221},
  archivePrefix={arXiv},
  primaryClass={cs.CL},
  url={https://arxiv.org/abs/2207.05221}
}

@inproceedings{kuhn2023semantic,
  title={Semantic Uncertainty: Linguistic Invariances for Uncertainty Estimation in Natural Language Generation},
  author={Kuhn, Lorenz and Gal, Yarin and Farquhar, Sebastian},
  booktitle={International Conference on Learning Representations},
  year={2023},
  url={https://openreview.net/forum?id=VD-AYtP0dve}
}

@inproceedings{kunitomo-jacquin-etal-2026-evidential,
  title={Evidential Semantic Entropy for {LLM} Uncertainty Quantification},
  author={Kunitomo-Jacquin, Lucie and Marrese-Taylor, Edison and Fukuda, Ken and Hamasaki, Masahiro},
  booktitle={Proceedings of the 19th Conference of the European Chapter of the Association for Computational Linguistics (Volume 1: Long Papers)},
  pages={7107--7122},
  year={2026},
  address={Rabat, Morocco},
  publisher={Association for Computational Linguistics},
  doi={10.18653/v1/2026.eacl-long.334},
  url={https://aclanthology.org/2026.eacl-long.334/}
}

@inproceedings{li-etal-2021-p,
  title={{P}-Stance: A Large Dataset for Stance Detection in Political Domain},
  author={Li, Yingjie and Sosea, Tiberiu and Sawant, Aditya and Nair, Ajith Jayaraman and Inkpen, Diana and Caragea, Cornelia},
  booktitle={Findings of ACL-IJCNLP 2021},
  year={2021},
  pages={2355--2365},
  publisher={Association for Computational Linguistics},
  doi={10.18653/v1/2021.findings-acl.208},
  url={https://aclanthology.org/2021.findings-acl.208/}
}

@inproceedings{li-etal-2026-march,
  title={{MARCH}: Multi-Agent Reinforced Check for Hallucination},
  author={Li, Zhuo and Zhang, Yupeng and Cheng, Pengyu and Song, Jiajun and Zhou, Mengyu and Li, Hao and Hu, Shujie and Qin, Yu and Zhao, Erchao and Jiang, Xiaoxi and Jiang, Guanjun},
  booktitle={Proceedings of the 64th Annual Meeting of the Association for Computational Linguistics (Volume 1: Long Papers)},
  year={2026},
  pages={39389--39415},
  publisher={Association for Computational Linguistics},
  url={https://aclanthology.org/2026.acl-long.1828/},
  doi={10.18653/v1/2026.acl-long.1828}
}

@inproceedings{malinin2018predictive,
  title={Predictive Uncertainty Estimation via Prior Networks},
  author={Malinin, Andrey and Gales, Mark},
  booktitle={Advances in Neural Information Processing Systems},
  volume={31},
  year={2018},
  url={https://papers.nips.cc/paper_files/paper/2018/hash/3ea2db50e62ceefceaf70a9d9a56a6f4-Abstract.html}
}

@inproceedings{manakul-etal-2023-selfcheckgpt,
  title={{SelfCheckGPT}: Zero-Resource Black-Box Hallucination Detection for Generative Large Language Models},
  author={Manakul, Potsawee and Liusie, Adian and Gales, Mark},
  booktitle={Proceedings of EMNLP 2023},
  year={2023},
  pages={9004--9017},
  doi={10.18653/v1/2023.emnlp-main.557},
  url={https://aclanthology.org/2023.emnlp-main.557/}
}

@misc{mistralai2025small2501,
  title={Mistral Small 24B Instruct 2501},
  author={{Mistral AI}},
  year={2025},
  howpublished={Hugging Face Model Card},
  url={https://huggingface.co/mistralai/Mistral-Small-24B-Instruct-2501}
}

@inproceedings{mohammad-etal-2016-semeval,
  title={{SemEval}-2016 {T}ask 6: Detecting Stance in Tweets},
  author={Mohammad, Saif and Kiritchenko, Svetlana and Sobhani, Parinaz and Zhu, Xiaodan and Cherry, Colin},
  booktitle={Proceedings of SemEval-2016},
  pages={31--41},
  year={2016},
  publisher={Association for Computational Linguistics},
  doi={10.18653/v1/S16-1003},
  url={https://aclanthology.org/S16-1003/}
}

@inproceedings{nguyen-etal-2025-beyond,
  title={Beyond Semantic Entropy: Boosting {LLM} Uncertainty Quantification with Pairwise Semantic Similarity},
  author={Nguyen, Dang and Payani, Ali and Mirzasoleiman, Baharan},
  booktitle={Findings of ACL 2025},
  year={2025},
  pages={4530--4540},
  doi={10.18653/v1/2025.findings-acl.234},
  url={https://aclanthology.org/2025.findings-acl.234/}
}

@inproceedings{nguyen-kim-2025-external,
  title={Is External Information Useful for Stance Detection with {LLM}s?},
  author={Nguyen, Quang Minh and Kim, Taegyoon},
  booktitle={Findings of ACL 2025},
  year={2025},
  pages={14798--14807},
  publisher={Association for Computational Linguistics},
  doi={10.18653/v1/2025.findings-acl.764},
  url={https://aclanthology.org/2025.findings-acl.764/}
}

@inproceedings{shelmanov2025tutorial,
  title={Uncertainty Quantification for Large Language Models},
  author={Shelmanov, Artem and Panov, Maxim and Vashurin, Roman and Vazhentsev, Artem and Fadeeva, Ekaterina and Baldwin, Timothy},
  booktitle={Proceedings of the 63rd Annual Meeting of the Association for Computational Linguistics (Volume 5: Tutorial Abstracts)},
  pages={3--4},
  year={2025},
  publisher={Association for Computational Linguistics},
  doi={10.18653/v1/2025.acl-tutorials.3},
  url={https://aclanthology.org/2025.acl-tutorials.3/}
}

@inproceedings{tan-etal-2025-consistent,
  title={Too Consistent to Detect: A Study of Self-Consistent Errors in {LLM}s},
  author={Tan, Hexiang and Sun, Fei and Liu, Sha and Su, Du and Cao, Qi and Chen, Xin and Wang, Jingang and Cai, Xunliang and Wang, Yuanzhuo and Shen, Huawei and Cheng, Xueqi},
  booktitle={Proceedings of EMNLP 2025},
  year={2025},
  pages={4755--4765},
  doi={10.18653/v1/2025.emnlp-main.238},
  url={https://aclanthology.org/2025.emnlp-main.238/}
}

@inproceedings{geifman2017selective,
  title={Selective Classification for Deep Neural Networks},
  author={Geifman, Yonatan and El-Yaniv, Ran},
  booktitle={Advances in Neural Information Processing Systems},
  volume={30},
  pages={4878--4887},
  year={2017},
  url={https://arxiv.org/abs/1705.08500}
}

@inproceedings{varshney2022selective,
  title={Towards Improving Selective Prediction Ability of {NLP} Systems},
  author={Varshney, Neeraj and Mishra, Swaroop and Baral, Chitta},
  booktitle={Proceedings of the 7th Workshop on Representation Learning for NLP},
  pages={221--226},
  year={2022},
  publisher={Association for Computational Linguistics},
  doi={10.18653/v1/2022.repl4nlp-1.23},
  url={https://aclanthology.org/2022.repl4nlp-1.23/}
}

@inproceedings{wastl-etal-2025-uzh,
  title={{UZH} at {S}em{E}val-2025 Task 3: Token-Level Self-Consistency for Hallucination Detection},
  author={Wastl, Michelle and Vamvas, Jannis and Sennrich, Rico},
  booktitle={Proceedings of the 19th International Workshop on Semantic Evaluation (SemEval-2025)},
  pages={257--270},
  year={2025},
  publisher={Association for Computational Linguistics},
  url={https://aclanthology.org/2025.semeval-1.38/}
}

@article{xue2025verify,
  title={Verify when Uncertain: Beyond Self-Consistency in Black Box Hallucination Detection},
  author={Xue, Yihao and Greenewald, Kristjan and Mroueh, Youssef and Mirzasoleiman, Baharan},
  journal={Transactions on Machine Learning Research},
  year={2026},
  url={https://arxiv.org/abs/2502.15845},
}

@inproceedings{yang-etal-2024-reinforcement,
  title={Reinforcement Tuning for Detecting Stances and Debunking Rumors Jointly with Large Language Models},
  author={Yang, Ruichao and Gao, Wei and Ma, Jing and Lin, Hongzhan and Wang, Bo},
  booktitle={Findings of the Association for Computational Linguistics: ACL 2024},
  year={2024},
  publisher={Association for Computational Linguistics},
  doi={10.18653/v1/2024.findings-acl.796},
  url={https://aclanthology.org/2024.findings-acl.796/},
  pages={13423--13439},
}

@inproceedings{zhang-etal-2023-cross-lingual,
  title={Cross-Lingual Cross-Target Stance Detection with Dual Knowledge Distillation Framework},
  author={Zhang, Ruike and Yang, Hanxuan and Mao, Wenji},
  booktitle={Proceedings of the 2023 Conference on Empirical Methods in Natural Language Processing},
  year={2023},
  publisher={Association for Computational Linguistics},
  doi={10.18653/v1/2023.emnlp-main.666},
  url={https://aclanthology.org/2023.emnlp-main.666/},
  pages={10804--10819},
  address={Singapore},
}

@inproceedings{zhang-etal-2023-sac3,
  title={{SAC}$^3$: Reliable Hallucination Detection in Black-Box Language Models via Semantic-aware Cross-check Consistency},
  author={Zhang, Jiaxin and Li, Zhuohang and Das, Kamalika and Malin, Bradley and Kumar, Sricharan},
  booktitle={Findings of EMNLP 2023},
  year={2023},
  pages={15445--15458},
  doi={10.18653/v1/2023.findings-emnlp.1032},
  url={https://aclanthology.org/2023.findings-emnlp.1032/}
}

@inproceedings{zhang-etal-2024-llm-driven,
  title={{LLM}-Driven Knowledge Injection Advances Zero-Shot and Cross-Target Stance Detection},
  author={Zhang, Zhao and Li, Yiming and Zhang, Jin and Xu, Hui},
  booktitle={Proceedings of NAACL 2024 (Short Papers)},
  year={2024},
  pages={371--378},
  publisher={Association for Computational Linguistics},
  doi={10.18653/v1/2024.naacl-short.32},
  url={https://aclanthology.org/2024.naacl-short.32/}
}

@inproceedings{zhang-etal-2025-bold,
  title={Bold Claims or Self-Doubt? Factuality Hallucination Type Detection via Belief State},
  author={Zhang, Dongyu and Hong, Qingqing and Hou, Bingxuan and Lin, Jiayi and Zhang, Chenyang and Li, Jialin and Wang, Junli},
  booktitle={Findings of the Association for Computational Linguistics: EMNLP 2025},
  year={2025},
  doi={10.18653/v1/2025.findings-emnlp.527},
  url={https://aclanthology.org/2025.findings-emnlp.527/},
  pages={9946--9959},
  address={Suzhou, China},
}

@inproceedings{zhang-etal-2025-mprf,
  title={{MPRF}: Interpretable Stance Detection through Multi-Path Reasoning Framework},
  author={Zhang, ZhaoDan and Zhang, Jin and Xu, Hui and Guo, Jiafeng and Cheng, Xueqi},
  booktitle={Proceedings of EMNLP 2025},
  year={2025},
  pages={454--470},
  publisher={Association for Computational Linguistics},
  doi={10.18653/v1/2025.emnlp-main.24},
  url={https://aclanthology.org/2025.emnlp-main.24/}
}

@inproceedings{zhang-etal-2025-mpvstance,
  title={{MPVStance}: Mitigating Hallucinations in Stance Detection with Multi-Perspective Verification},
  author={Zhang, ZhaoDan and Zhang, Zhao and Zhang, Jin and Xu, Hui and Cheng, Xueqi},
  booktitle={Proceedings of ACL 2025},
  year={2025},
  pages={1053--1067},
  publisher={Association for Computational Linguistics},
  doi={10.18653/v1/2025.acl-long.53},
  url={https://aclanthology.org/2025.acl-long.53/}
}

@inproceedings{zhang2026cognitive,
  title={Modeling Human-Like Cognition for Stance Detection: Integrating Intuitive Judgment and Analytical Reasoning},
  author={Zhang, Zhaodan and Zhang, Jin and Guo, Jiafeng and Cheng, Xueqi},
  booktitle={Proceedings of the 64th Annual Meeting of the Association for Computational Linguistics (Volume 1: Long Papers)},
  pages={5852--5867},
  year={2026},
  publisher={Association for Computational Linguistics},
  doi={10.18653/v1/2026.acl-long.264},
  url={https://aclanthology.org/2026.acl-long.264/},
  address={San Diego, California, United States},
}

@misc{meta2024llama33,
  author       = {{Meta AI}},
  title        = {{Llama-3.3-70B-Instruct}},
  year         = {2024},
  howpublished = {Hugging Face Model Card},
  url          = {https://huggingface.co/meta-llama/Llama-3.3-70B-Instruct},
  note         = {Released December 6, 2024}
}

@misc{sabbaghan2026moprosd,
      title={MoPrO-SD: Modular Prompt Optimization for Stance Detection}, 
      author={Meysam Sabbaghan and Nikesh Gyawali and Arman Zareian Jahromi and Sebastian Loera-Lozano and Doina Caragea and Cornelia Caragea},
      year={2026},
      eprint={2610.13386},
      archivePrefix={arXiv},
      primaryClass={cs.CL},
      url={https://arxiv.org/abs/2610.13386}, 
}

\clearpage
\appendix

\appendix

% Reset appendix figure and table numbering
\setcounter{figure}{0}
\setcounter{table}{0}

% Appendix figures: A1, A2, A3, ...
\renewcommand{\thefigure}{A\arabic{figure}}

% Appendix tables: T1, T2, T3, ...
\renewcommand{\thetable}{T\arabic{table}}

\section{Dataset Statistics}
\label{app:dataset_statistics}

This appendix summarizes the evaluation data used in
B-CARE. Tables~\ref{tab:dataset_statistics} and
\ref{tab:target_distribution} report dataset-level
and target-wise statistics, respectively.
The evaluation includes 5,206 examples across
13 targets and four benchmarks, comprising three
ternary stance datasets and one binary dataset.

\begin{table}[htbp]
\centering
\small
\setlength{\tabcolsep}{7pt}
\begin{tabular}{lrrrrr}
\toprule
\textbf{Dataset} & $\boldsymbol{N}$ &
\textbf{Tgt.} & \textbf{F} &
\textbf{A} & \textbf{N} \\
\midrule
SemEval  & 1249 & 5 & 304 & 715 & 230 \\
Vaccines & 1000 & 1 & 396 & 259 & 345 \\
COVID    & 800  & 4 & 263 & 243 & 294 \\
P-Stance & 2157 & 3 & 1032 & 1125 & -- \\
\midrule
\textbf{Total}
& \textbf{5206} & \textbf{13}
& \textbf{1995} & \textbf{2342} & \textbf{869} \\
\bottomrule
\end{tabular}
\caption{Evaluation-set statistics.
$N$ denotes the number of unique test examples,
and Tgt. denotes the number of targets.
F, A, and N represent FAVOR, AGAINST, and NONE,
respectively. P-Stance does not contain
the NONE class.}
\label{tab:dataset_statistics}
\end{table}

\begin{table}[htbp]
\centering
\small
\setlength{\tabcolsep}{4pt}
\begin{tabular}{llr}
\toprule
\textbf{Dataset} & \textbf{Target} &
\textbf{Test} \\
\midrule
\multirow{5}{*}{SemEval-2016}
& Atheism & 220 \\
& Climate Change & 169 \\
& Feminist Movement & 285 \\
& Hillary Clinton & 295 \\
& Legalization of Abortion & 280 \\
\cmidrule(lr){2-3}
& \textbf{Subtotal} & \textbf{1249} \\
\midrule
VaccineStance
& Vaccines (single target) & 1000 \\
\midrule
\multirow{4}{*}{COVID-19-Stance}
& Anthony Fauci & 200 \\
& Face Masks & 200 \\
& School Closures & 200 \\
& Stay at Home Orders & 200 \\
\cmidrule(lr){2-3}
& \textbf{Subtotal} & \textbf{800} \\
\midrule
\multirow{3}{*}{P-Stance}
& Donald Trump & 777 \\
& Joe Biden & 745 \\
& Bernie Sanders & 635 \\
\cmidrule(lr){2-3}
& \textbf{Subtotal} & \textbf{2157} \\
\midrule
\multicolumn{2}{l}{\textbf{Total (13 targets)}}
& \textbf{5206} \\
\bottomrule
\end{tabular}
\caption{Target-wise distribution of test examples
across the four evaluation benchmarks.
VaccineStance contains a single target (Vaccines),
whereas SemEval-2016, COVID-19-Stance, and P-Stance
contain five, four, and three targets, respectively.
Counts correspond to the evaluation sets used in B-CARE.}
\label{tab:target_distribution}
\end{table}

\section{Verifier Prompt Templates}
\label{app:verifier_prompts}

We provide the exact verifier prompts used in the
B-CARE experiments and the principal verification
controls. B-CARE evaluates classwise textual evidence
without access to the base model's prediction,
confidence, or gold label. The three-class template
is used for SemEval-2016, VaccineStance, and
COVID-19-Stance, while P-Stance uses a binary
variant. We additionally document the
prediction-aware PACE modification and the
direct-correctness EDCV control.

\subsection{B-CARE Prediction-Blind Evidence Prompts}

\begin{bcareprompt}{A1. B-CARE Three-Class Evidence Prompt}

\promptheading{Input}

\texttt{TARGET:}

\promptfield{{target}}

\medskip
\texttt{TEXT:}

\promptfield{{text}}

\promptheading{Instructions}

Evaluate the textual evidence for each possible
stance of the TEXT toward the TARGET.

Do not decide which stance label is correct overall.

Instead, evaluate each hypothesis independently
based only on evidence contained or clearly implied
in the TEXT.

\promptheading{FAVOR}

Is there explicit or clearly implied evidence that
the author supports, endorses, agrees with, or
expresses a favorable position toward the TARGET?

\promptheading{AGAINST}

Is there explicit or clearly implied evidence that
the author opposes, rejects, criticizes, disagrees
with, or expresses an unfavorable position toward
the TARGET?

\promptheading{NONE}
Is there explicit or clearly implied evidence that the author does not express a discernible stance of support or opposition toward the TARGET, such as presenting factual information, asking a neutral question, or discussing the topic without taking a position?

\promptheading{Evidence-Strength Scale}

For each hypothesis return an evidence-strength
score:

\begin{tabular}{@{}cl@{}}
\texttt{0} & no supporting evidence\\
\texttt{1} & weak or ambiguous evidence\\
\texttt{2} & moderate evidence\\
\texttt{3} & strong and direct evidence
\end{tabular}

\promptheading{Output Format}

Return JSON only:

\begin{Verbatim}[fontsize=\scriptsize,
                 breaklines=true]
{
  "FAVOR": 0,
  "AGAINST": 0,
  "NONE": 0
}
\end{Verbatim}

Do not output a final stance label.

Do not explain your answer.

\end{bcareprompt}

\begin{bcareprompt}{A2. B-CARE Binary Evidence Prompt (P-Stance)}

\promptheading{Input}

\texttt{TARGET:}

\promptfield{{target}}

\medskip
\texttt{TEXT:}

\promptfield{{text}}

\promptheading{Instructions}

Evaluate the textual evidence for each possible
stance of the TEXT toward the TARGET.

Do not decide which stance label is correct overall.

Instead, evaluate each hypothesis independently
based only on evidence contained or clearly implied
in the TEXT.

\promptheading{FAVOR}

Is there explicit or clearly implied evidence that
the author supports, endorses, agrees with, or
expresses a favorable position toward the TARGET?

\promptheading{AGAINST}

Is there explicit or clearly implied evidence that
the author opposes, rejects, criticizes, disagrees
with, or expresses an unfavorable position toward
the TARGET?

\promptheading{Evidence-Strength Scale}

For each hypothesis return an evidence-strength
score:

\begin{tabular}{@{}cl@{}}
\texttt{0} & no supporting evidence\\
\texttt{1} & weak or ambiguous evidence\\
\texttt{2} & moderate evidence\\
\texttt{3} & strong and direct evidence
\end{tabular}

\promptheading{Output Format}

Return JSON only:

\begin{Verbatim}[fontsize=\scriptsize,
                 breaklines=true]
{
  "FAVOR": 0,
  "AGAINST": 0
}
\end{Verbatim}

Do not output a final stance label.

Do not explain your answer.

\end{bcareprompt}

\subsection{Prediction-Aware Control}

\begin{bcareprompt}{A3. PACE Prediction-Aware Modification}

PACE uses the corresponding B-CARE evidence
template, with the following additional field
inserted immediately after the TEXT input:

\medskip

\begin{Verbatim}[fontsize=\scriptsize]
BASE PREDICTION:
{base_prediction}
\end{Verbatim}

\medskip

All remaining instructions, evidence-strength
definitions, and output fields are unchanged.

For three-class datasets, PACE uses the
three-class B-CARE template (A1).

For P-Stance, PACE uses the binary B-CARE
template (A2).

\end{bcareprompt}

\subsection{Direct Correctness Verification Control}

\begin{bcareprompt}{A4. EDCV Three-Class Prompt}

\promptheading{System Message}

Respond with exactly one integer character:
0, 1, 2, or 3. Do not write any other text.

\promptheading{User Message}

Task: Evaluate whether a stance prediction
is correct.

\medskip
Target:

\promptfield{{target}}

\medskip
Text:

\promptfield{{text}}

\medskip
Predicted stance:

\promptfield{{base_prediction}}

\medskip
The possible stance labels are FAVOR, AGAINST,
and NONE.

Determine how strongly the text supports the
claim that the predicted stance is correct.

Return an integer correctness score from 0 to 3:

\medskip
0 = the predicted stance is clearly incorrect

1 = the predicted stance is probably incorrect

2 = the predicted stance is probably correct

3 = the predicted stance is clearly correct

\medskip
Output exactly one integer: 0, 1, 2, or 3.

\end{bcareprompt}

\begin{bcareprompt}{A5. EDCV Binary Prompt (P-Stance)}

\promptheading{System Message}

Respond with exactly one integer character:
0, 1, 2, or 3. Do not write any other text.

\promptheading{User Message}

Task: Evaluate whether a stance prediction
is correct.

\medskip
Target:

\promptfield{{target}}

\medskip
Text:

\promptfield{{text}}

\medskip
Predicted stance:

\promptfield{{base_prediction}}

\medskip
The possible stance labels are FAVOR and AGAINST.

Determine how strongly the text supports the
claim that the predicted stance is correct.

Return an integer correctness score from 0 to 3:

\medskip
0 = the predicted stance is clearly incorrect

1 = the predicted stance is probably incorrect

2 = the predicted stance is probably correct

3 = the predicted stance is clearly correct

\medskip
Output exactly one integer: 0, 1, 2, or 3.

\end{bcareprompt}

\section{Baseline and Control Implementation Details}
\label{app:baseline_implementation}
%\label{app:baselines}

All methods assign higher risk scores to instances more
likely to contain incorrect base-model predictions.
Unless stated otherwise, $\hat{y}$ denotes the frozen
base-model prediction.

The three base models (Phi, Mistral, and Gemma) use
greedy decoding (\texttt{do\_sample=false},
\texttt{max\_new\_tokens=16}), without an explicit
temperature passed to generation. The fixed external
Llama verifier uses temperature $0.0$ and
\texttt{max\_tokens=80}.

The sampling-based uncertainty baselines (LE, VR,
SE, and EVSE) use $M=5$ generations, temperature
$1.0$, top-$p=0.9$, and
\texttt{max\_new\_tokens=8}. MSP, PE, and DSV
use token-probability scoring rather than sampling.
The $p(\mathrm{True})$ baseline also uses
token-probability scoring for its final risk score,
with temperature $1.0$ used to generate brainstorm
responses.

\paragraph{Maximum Softmax Probability and Predictive Entropy.}
For each stance class $c$, we compute the sum of
teacher-forced token log-probabilities of its label
verbalizer, without length normalization. Applying
softmax to these scores gives $p(c\mid x)$.
Following \citet{hendrycks2017baseline}, the
error-risk scores are
\begin{align}
R_{\mathrm{MSP}}
&=1-\max_c p(c\mid x),\\
R_{\mathrm{PE}}
&=-\sum_c p(c\mid x)\log p(c\mid x).
\end{align}
Both scores are computed from the frozen base model.

\paragraph{Prediction Alignment for MSP and PE.}
MSP and Predictive Entropy use the teacher-forced
class-verbalizer distribution rather than the frozen
greedy-decoded label directly. To assess alignment,
we compared the greedy-decoded prediction with the
argmax of the verbalizer probabilities. Across all
12 dataset--model settings, 20 mismatches were
observed among 15,618 test examples (0.128\%).
Thus, the two prediction procedures agree for
nearly all evaluated examples, although MSP and PE
quantify uncertainty under the verbalizer
distribution rather than confidence in the frozen
greedy prediction.

\paragraph{$p(\mathrm{True})$.}
We adapt Kadavath-style self-evaluation using the
open Semantic Uncertainty implementation
\cite{kadavath2022language,farquhar2024detecting}.
The candidate answer is the frozen base prediction
$\hat{y}$. The procedure uses 20 TRAIN-only
few-shot demonstrations (seed 2026092817) and
ten brainstormed stance responses. No TEST gold
labels are used in prompt construction.

Following the reference implementation's
\texttt{p\_false\_fixed} convention, error risk is
\begin{equation}
R_{p(\mathrm{True})}
=1-P(\texttt{ A}\mid q),
\end{equation}
where $q$ is the complete A/B self-evaluation
prompt. We extract the raw-completion token
probability without chat-template wrapping or
renormalization over A/B alternatives. All reported
results use the corrected implementation.

\paragraph{Direct Self-Verification.}
DSV is a separate direct-correctness control.
The base model assesses whether its predicted
stance is correct using a True/False prompt.
Unlike $p(\mathrm{True})$, DSV normalizes the
two answer probabilities:
\begin{equation}
\begin{aligned}
R_{\mathrm{DSV}}
&=1-P(\mathrm{True}),\\
P(\mathrm{True})
&=\frac{e^{\ell_{\mathrm{True}}}}
{e^{\ell_{\mathrm{True}}}
+e^{\ell_{\mathrm{False}}}}.
\end{aligned}
\end{equation}
Here, $\ell_{\mathrm{True}}$ and
$\ell_{\mathrm{False}}$ are the corresponding
token log-probabilities.

\paragraph{Label Entropy and Variation Ratio.}
Let $n_c$ be the number of valid sampled responses
assigned to class $c$, and let
$n_{\mathrm{valid}}=\sum_c n_c$.
Label Entropy (LE) and Variation Ratio (VR) are
\begin{align}
R_{\mathrm{LE}}
&=-\sum_c \frac{n_c}{n_{\mathrm{valid}}}
\log\frac{n_c}{n_{\mathrm{valid}}},\\
R_{\mathrm{VR}}
&=1-\frac{\max_c n_c}{M}.
\end{align}
LE uses valid parsed labels, whereas VR retains
the planned sample count $M$ in its denominator.

\paragraph{Semantic Entropy.}
Semantic Entropy (SE)
\cite{kuhn2023semantic,farquhar2024detecting}
uses $M=5$ stochastic generations from the base stance
prompt. Generated responses are mapped to canonical
stance labels when parsing succeeds; otherwise, the
original decoded response is retained, with empty
responses represented explicitly. No response is
excluded from SE clustering solely because it cannot
be parsed as a stance label.

Semantic relationships are evaluated using
DeBERTa-v2-xlarge-MNLI
(\url{https://huggingface.co/microsoft/deberta-v2-xlarge-mnli}),
with the input text and target included in the NLI
context. Bidirectional NLI judgments group
semantically equivalent responses. Sample sequence
likelihoods are computed from token log-probabilities,
normalized across the sampled responses, and summed
within semantic clusters. The SE risk score is
\begin{equation}
R_{\mathrm{SE}}=-\sum_k e^{\ell_k}\ell_k,
\end{equation}
where $\ell_k$ denotes the normalized log-probability
mass of semantic cluster $k$.

\paragraph{Evidential Semantic Entropy.}
Evidential Semantic Entropy (EVSE)
\cite{kunitomo-jacquin-etal-2026-evidential}
uses the same sampled and parsed response set and
NLI model, but organizes pairwise entailment
relationships into a hierarchy. Entailment cycles
are merged, and the resulting structure includes
nodes representing unresolved or uncommitted evidence.
Probability masses are assigned from the sequence
likelihoods of unique observed responses; residual
mass is allocated to the ignorance set. The resulting
basic belief assignment is transformed into a
pignistic probability distribution, whose entropy
provides the reported uncertainty score.

\paragraph{Same-Model Comparative Evidence.}
This control replaces the fixed external verifier
with the corresponding base model. The base model
generates prediction-blind classwise ordinal
evidence using the same evidence scale, prompt,
and output schema as B-CARE. Local generation
uses greedy decoding (\texttt{do\_sample=false},
\texttt{max\_new\_tokens=96}); the OpenRouter
same-model path uses temperature $0.0$ and
\texttt{max\_tokens=80}. The risk remains
\begin{equation}
R_{\mathrm{same}}=A_B-S_B.
\end{equation}
This comparison preserves the evidence rubric,
prediction-blind prompting, and relative-evidence
ranking rule while changing the verifier model.
However, differences in model architecture,
inference backend, generation settings, and
chat formatting prevent attributing the observed
performance differences exclusively to verifier
capacity. The controlled ablations of prediction
blindness and graded versus binary evidence
provide complementary evidence for the
contributions of B-CARE's design.

\paragraph{Evidence-Label Disagreement.}
Using the frozen external-verifier evidence
vector $\mathbf{E}$, we define
\begin{equation}
R_{\mathrm{LD}}=
\begin{cases}
0, & E_{\hat{y}}>\max_{c\ne\hat{y}}E_c,\\
1, & \text{otherwise}.
\end{cases}
\end{equation}
Ties at the maximum evidence level count as
disagreement. Unlike B-CARE, this binary control
discards the magnitude of comparative evidence.

\paragraph{Cross-Base Disagreement.}
Let $\hat{y}$ be the evaluated base prediction,
and let $o_1,o_2$ be the predictions of the other
two base models. Cross-Base Disagreement (CBD)
assigns the following ordinal risk:
\begin{equation}
\begingroup
\small
R_{\mathrm{CBD}}=
\begin{cases}
0, & o_1=o_2=\hat{y},\\
1, & \sum_{i=1}^{2}\mathbb{I}[o_i=\hat{y}]=1,\\
2, & o_1=o_2\ne\hat{y},\\
3, & |\{\hat{y},o_1,o_2\}|=3.
\end{cases}
\endgroup
\end{equation}
Here, $\mathbb{I}[\cdot]$ is the indicator
function. CBD uses only hard predictions,
without probability or evidence scores.

\paragraph{External Direct Correctness Verification.}
EDCV provides the external verifier with the
target, input text, and base prediction $\hat{y}$.
The verifier returns a graded correctness
assessment $C\in\{0,1,2,3\}$, yielding
\begin{equation}
R_{\mathrm{EDCV}}=3-C.
\end{equation}
EDCV uses the external Llama verifier with
temperature $0.0$ and \texttt{max\_tokens=16}.
Unlike B-CARE, EDCV explicitly conditions on
the base prediction and assesses its correctness
directly.

\paragraph{Prediction-Aware Comparative Evidence.}
PACE retains the external verifier, classwise
evidence scale, output schema, and comparative
risk $A_B-S_B$, but additionally reveals the
base prediction $\hat{y}$ in the verifier prompt.
It uses temperature $0.0$ and
\texttt{max\_tokens=80}. Thus, the intended
difference from prediction-blind B-CARE is
exposure of $\hat{y}$ before evidence generation.

\paragraph{Evidence Components and Alternative Rankings.}
For frozen evidence $\mathbf{E}$, let
$S_B=E_{\hat{y}}$ and
$A_B=\max_{c\ne\hat{y}}E_c$.
We compare the primary score $A_B-S_B$ with
support-only $-S_B$ and alternative-only $A_B$.

As an additional ranking sensitivity analysis,
we consider normalized evidence share:
\begin{equation}
R_{\mathrm{share}}
=1-\frac{E_{\hat{y}}}{\sum_c E_c}.
\end{equation}
When $\sum_c E_c=0$, normalized evidence share
uses the uniform class share.

\paragraph{Shared Evaluation Conventions.}
AUROC calculations use midranks for tied risk
scores. ErrorCapture uses fractional allocation
within tied risk groups at the selection boundary.
Paired bootstrap comparisons use 10,000 resamples
unless a method-specific protocol specifies
otherwise.

\paragraph{Tie Handling in AURC and E-AURC.}
For risk--coverage evaluation, instances are ranked
in ascending order of estimated error risk. When
multiple instances receive identical risk scores,
ties are handled using expected fractional allocation.
Specifically, for a group of $g$ tied instances
containing $e$ errors, each successive coverage step
receives an expected error contribution of $e/g$.
AURC is computed by averaging the selective error
rate across all coverage levels. E-AURC is obtained
by subtracting the oracle AURC, which ranks all
correct predictions before incorrect predictions.
This treatment avoids dependence on arbitrary
ordering within tied groups.

\subsection{Additional Error-Ranking Controls}
\label{app:additional_controls}

We evaluate four complementary controls
to further examine the contributions of
classwise evidence, graded scoring, and
chance-level ranking. All comparisons
cover the complete set of 12 dataset--model
settings across four benchmarks and three
base models.

\paragraph{Same-Model Label Disagreement.}
This control uses prediction-blind ordinal
classwise evidence generated by the
corresponding base model. An instance is
assigned risk zero only when its frozen
base prediction uniquely maximizes the
classwise evidence. All other cases,
including evidence ties, receive risk one.
Unlike Same-Model Comparative Evidence,
which retains the graded relative-evidence
score, this control discards evidence
magnitude and uses binary disagreement.

\paragraph{Llama Label Disagreement.}
This control compares the frozen base-model
stance prediction with a separately generated
stance label from the external Llama verifier.
Unlike Evidence-Label Disagreement, which uses
the ordinal classwise evidence vector and a
unique-maximum rule, this control compares
two discrete stance labels directly.

\paragraph{Random Ranking.}
We assign independent risk scores sampled
from a uniform distribution and calculate
the mean error-ranking AUROC across 50
fixed random seeds. This provides a
chance-level reference under the same
frozen test predictions and error labels.

\paragraph{Normalized Evidence Share.}
This control uses the same prediction-blind
classwise evidence as B-CARE but replaces
the original comparative score with
\begin{equation}
R_{\mathrm{share}}(B,x)
=
1-
\frac{E_{\hat{y}_B}(x)}
{\sum_{c \in \mathcal{Y}} E_c(x)}.
\end{equation}
When the total evidence is zero, the
predicted-class evidence share is defined
as $1/|\mathcal{Y}|$. This variant tests
whether normalized support for the
predicted class provides an effective
alternative to the original $A_B-S_B$
ranking score.

\begin{table}[t]
\centering
\small
\setlength{\tabcolsep}{4pt}
\begin{tabular}{lccc}
\toprule
\textbf{Control} &
\textbf{Settings} &
\textbf{AUROC} &
$\boldsymbol{\Delta}$ \textbf{vs. B-CARE} \\
\midrule
Same-Model LD & 12 & 0.5856 & +0.1490 \\
Llama LD & 12 & 0.6704 & +0.0642 \\
Random Ranking & 12 & 0.5008 & +0.2338 \\
$R_{\mathrm{share}}$ & 12 & 0.6940 & +0.0406 \\
\bottomrule
\end{tabular}
\caption{
Additional error-ranking controls across
12 dataset--model settings.
AUROC values are macro-averaged equally
across the 12 settings.
$\Delta$ denotes the difference between
the overall B-CARE mean AUROC (0.7346)
and the corresponding control mean AUROC.
Random Ranking averages 50 fixed-seed runs.
}
\label{tab:additional_controls}
\end{table}

B-CARE achieves higher mean AUROC than
all four controls across the 12 evaluated
dataset--model settings.

Compared with Same-Model LD, B-CARE
achieves a mean AUROC improvement of
0.1490 (0.7346 versus 0.5856).
This result indicates that B-CARE's
graded external comparative evidence
provides a stronger error-ranking signal
than binary disagreement derived from
the corresponding base model under the
evaluated conditions. However, this
comparison also changes verifier identity
and inference environment and therefore
does not isolate the contribution of
evidence granularity alone.

Compared with Llama LD, B-CARE achieves
a mean AUROC improvement of 0.0642
(0.7346 versus 0.6704).
Llama LD uses a separately generated
external Llama stance label rather than
graded classwise evidence. Consequently,
this comparison changes the form of
the verifier output and does not isolate
evidence-score granularity alone.

Random Ranking achieves approximately
chance-level performance, with a mean
AUROC of 0.5008. B-CARE improves upon
this control by 0.2338.

The normalized evidence-share variant
achieves a mean AUROC of 0.6940,
compared with 0.7346 for B-CARE,
corresponding to an improvement of
0.0406. Although B-CARE performs
better on average, normalized
evidence-share ranking can perform
better in individual settings,
indicating that the advantage of
the original comparative score
is setting-dependent.

\section{Statistical Significance Analysis}
\label{app:statistical_significance}

We assess the statistical significance of AUROC
differences using an example-level paired
nonparametric bootstrap within each dataset--model
setting. For each baseline, we compute
$\Delta\mathrm{AUROC} =
\mathrm{AUROC}(\textsc{B-CARE}) -
\mathrm{AUROC}(\mathrm{baseline})$
and obtain percentile 95\% confidence intervals
from bootstrap resampling (primarily
$B=10{,}000$). A significant win or loss is
recorded when the confidence interval lies
entirely above or below zero, respectively;
otherwise, the difference is classified as
nonsignificant. No correction for multiple
comparisons is applied.

Table~\ref{tab:bootstrap_full} summarizes the
significance outcomes across all 156 comparisons.
B-CARE achieves 122 significant wins, 11
significant losses, and 23 nonsignificant
differences. All significant losses occur
against confidence-based or self-verification
baselines: MSP, Predictive Entropy,
$p(\mathrm{True})$, and Self-Verification.
In contrast, no significant losses are
observed against the remaining nine baselines.
These results indicate broad improvements
over evidence-based alternatives, while
highlighting the complementary strengths
of confidence-based error ranking.
Nonsignificant differences should not be
interpreted as evidence of equivalence.

\begin{table}[H]
\centering
\footnotesize
\setlength{\tabcolsep}{3.5pt}
\renewcommand{\arraystretch}{1.10}
\begin{tabular}{lccc}
\toprule
\textbf{Baseline} &
\textbf{Win} &
\textbf{Loss} &
\textbf{NS} \\
\midrule
MSP                       & 4  & 4 & 4 \\
Predictive Entropy        & 4  & 4 & 4 \\
$p(\mathrm{True})$        & 8  & 2 & 2 \\
Self-Verification         & 9  & 1 & 2 \\
\midrule
Direct Verification       & 12 & 0 & 0 \\
Prediction-Aware Evidence & 7  & 0 & 5 \\
Same-Model Evidence       & 11 & 0 & 1 \\
Evidential SE             & 12 & 0 & 0 \\
Semantic Entropy          & 11 & 0 & 1 \\
Label Entropy             & 12 & 0 & 0 \\
Variation Ratio           & 12 & 0 & 0 \\
Cross-Model Disagreement  & 9  & 0 & 3 \\
Evidence-Label Disagreement     & 11 & 0 & 1 \\
\midrule
\textbf{Total} &
\textbf{122} &
\textbf{11} &
\textbf{23} \\
\bottomrule
\end{tabular}
\caption{
Summary of 156 paired bootstrap AUROC
comparisons across 12 dataset--model settings.
Win and Loss indicate significant positive
and negative differences, respectively;
NS indicates a nonsignificant difference
under unadjusted 95\% confidence intervals.
}
\label{tab:bootstrap_full}
\end{table}

\section{Complementarity Analysis}
\label{app:complementarity}

\begin{figure}[t]
\centering
\includegraphics[width=\columnwidth]%
{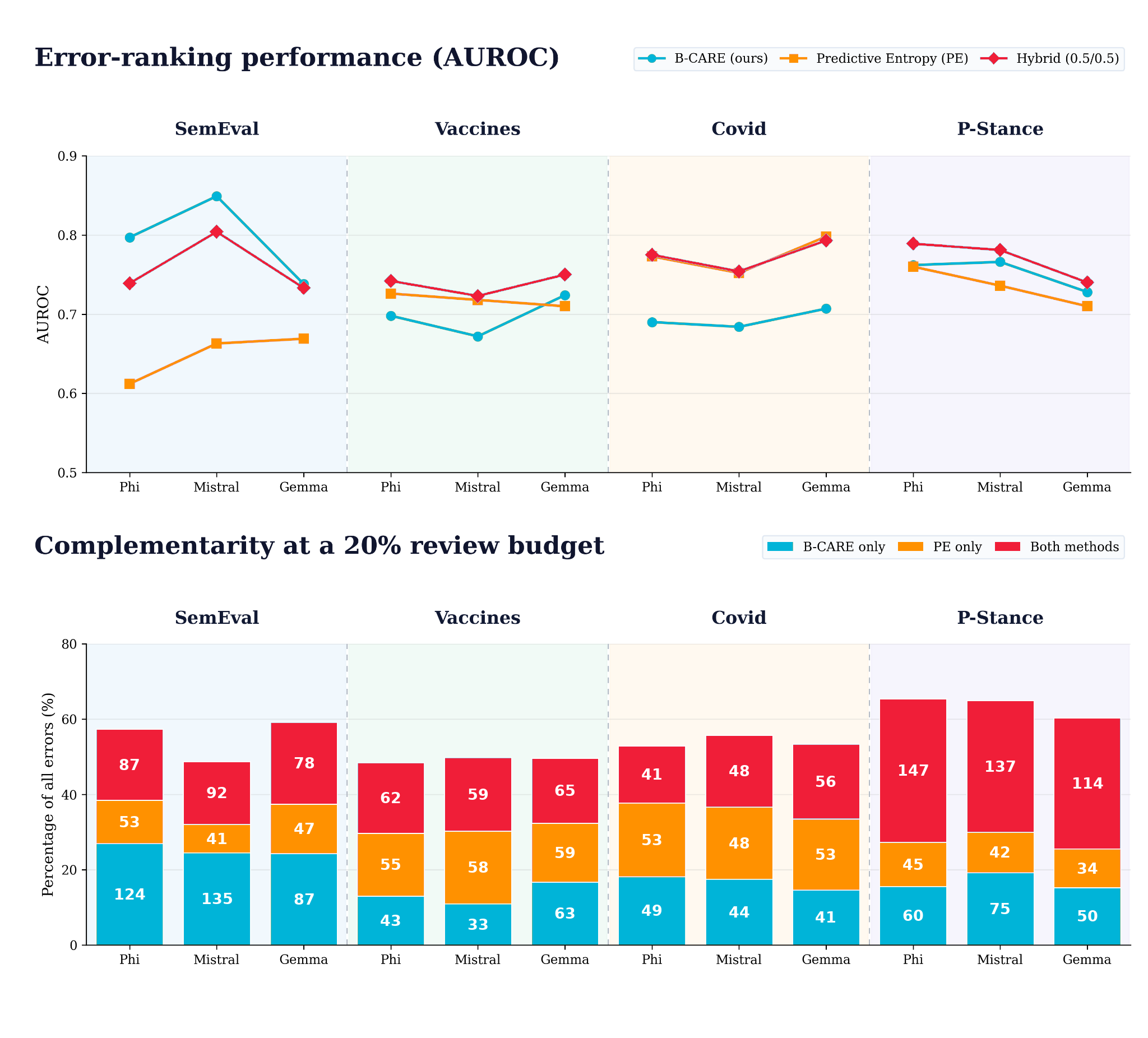}
\caption{
Complementarity between B-CARE and Predictive
Entropy (PE) across 12 dataset--model settings.
\textbf{Top:} AUROC of B-CARE, PE, and the
exploratory equal-weight hybrid.
\textbf{Bottom:} Errors captured exclusively
by B-CARE, exclusively by PE, or by both
within separate top-20\% review sets.
Segment heights represent percentages of
all errors; labels indicate error counts.
}
\label{fig:complementarity}
\end{figure}

We provide the per-setting results underlying
the complementarity analysis in
Section~\ref{sec:complementarity}.
For each dataset--model setting, B-CARE and
Predictive Entropy (PE) independently rank
test instances by estimated error risk.
At a 20\% review budget, we compare the
sets of actual errors identified by their
respective top-ranked instances.

We partition captured errors into three
mutually exclusive groups: errors identified
only by B-CARE, only by PE, and by both methods.
The union capture rate is the number of
distinct errors identified by either method
divided by the total number of actual errors.
Jaccard overlap is computed over the two
sets of captured errors. Because the union
combines two independently selected review
sets, it does not represent performance
under a single 20\% review budget.

To examine whether the two ranking signals
can be combined, we also evaluate an
exploratory fixed hybrid formed by averaging
their within-setting normalized ranks with
equal weights. No fusion weights are tuned.
This hybrid is a post-hoc diagnostic and is
not part of the frozen B-CARE method.

\begin{table*}[t]
\centering
\footnotesize
\setlength{\tabcolsep}{5pt}
\renewcommand{\arraystretch}{1.05}
\begin{tabular}{llccccc}
\toprule
\textbf{Dataset} &
\textbf{Model} &
\textbf{B-CARE} &
\textbf{PE} &
\textbf{Hybrid} &
$\boldsymbol{\Delta_{\mathrm{B}}}$ &
$\boldsymbol{\Delta_{\mathrm{PE}}}$ \\
\midrule
\multirow{3}{*}{SemEval}
& Phi     & \textbf{0.797} & 0.612 & 0.739 & $-0.058$ & $+0.127$ \\
& Mistral & \textbf{0.849} & 0.663 & 0.804 & $-0.045$ & $+0.141$ \\
& Gemma   & \textbf{0.738} & 0.669 & 0.733 & $-0.005$ & $+0.064$ \\
\midrule
\multirow{3}{*}{Vaccines}
& Phi     & 0.698 & 0.726 & \textbf{0.742} & $+0.044$ & $+0.017$ \\
& Mistral & 0.672 & 0.718 & \textbf{0.723} & $+0.051$ & $+0.005$ \\
& Gemma   & 0.724 & 0.710 & \textbf{0.750} & $+0.026$ & $+0.040$ \\
\midrule
\multirow{3}{*}{COVID}
& Phi     & 0.690 & 0.773 & \textbf{0.775} & $+0.085$ & $+0.002$ \\
& Mistral & 0.684 & 0.752 & \textbf{0.754} & $+0.070$ & $+0.002$ \\
& Gemma   & 0.707 & \textbf{0.798} & 0.793 & $+0.086$ & $-0.005$ \\
\midrule
\multirow{3}{*}{P-Stance}
& Phi     & 0.762 & 0.760 & \textbf{0.789} & $+0.027$ & $+0.028$ \\
& Mistral & 0.766 & 0.736 & \textbf{0.781} & $+0.016$ & $+0.045$ \\
& Gemma   & 0.728 & 0.710 & \textbf{0.740} & $+0.012$ & $+0.030$ \\
\midrule
\textbf{Macro mean}
& -- & 0.735 & 0.719 & \textbf{0.760}
& $+0.026$ & $+0.041$ \\
\bottomrule
\end{tabular}
\caption{
Per-setting AUROC of B-CARE, Predictive
Entropy (PE), and the exploratory fixed
equal-weight hybrid.
$\Delta_{\mathrm{B}}$ and
$\Delta_{\mathrm{PE}}$ denote hybrid AUROC
minus B-CARE and PE AUROC, respectively.
Bold indicates the highest AUROC in each
setting. Values are rounded to three decimals.
}
\label{tab:complementarity_auroc}
\end{table*}

\begin{table*}[t]
\centering
\small
\setlength{\tabcolsep}{6pt}
\renewcommand{\arraystretch}{1.10}
\begin{tabular}{llrrrrrrr}
\toprule
\textbf{Dataset} &
\textbf{Model} &
$\boldsymbol{k}$ &
\textbf{Errors} &
\textbf{B-only} &
\textbf{PE-only} &
\textbf{Both} &
\textbf{Union (\%)} &
\textbf{Jaccard} \\
\midrule
\multirow{3}{*}{SemEval}
& Phi     & 250 & 460 & 124 & 53 & 87 & 57.4 & 0.330 \\
& Mistral & 250 & 550 & 135 & 41 & 92 & 48.7 & 0.343 \\
& Gemma   & 250 & 358 & 87 & 47 & 78 & 59.2 & 0.368 \\
\midrule
\multirow{3}{*}{Vaccines}
& Phi     & 200 & 330 & 43 & 55 & 62 & 48.5 & 0.388 \\
& Mistral & 200 & 301 & 33 & 58 & 59 & 49.8 & 0.393 \\
& Gemma   & 200 & 377 & 63 & 59 & 65 & 49.6 & 0.348 \\
\midrule
\multirow{3}{*}{COVID}
& Phi     & 160 & 270 & 49 & 53 & 41 & 53.0 & 0.287 \\
& Mistral & 160 & 251 & 44 & 48 & 48 & 55.8 & 0.343 \\
& Gemma   & 160 & 281 & 41 & 53 & 56 & 53.4 & 0.373 \\
\midrule
\multirow{3}{*}{P-Stance}
& Phi     & 431 & 385 & 60 & 45 & 147 & 65.5 & 0.583 \\
& Mistral & 431 & 391 & 75 & 42 & 137 & 65.0 & 0.539 \\
& Gemma   & 431 & 328 & 50 & 34 & 114 & 60.4 & 0.576 \\
\midrule
\multicolumn{7}{l}{\textbf{Macro mean}} &
\textbf{55.5} & \textbf{0.406} \\
\bottomrule
\end{tabular}
\caption{
Overlap of actual errors captured by B-CARE
and Predictive Entropy (PE) within separately
selected top-20\% review sets.
$k$ is the number of reviewed instances
per method, and Errors is the total number
of incorrect base-model predictions.
B-only, PE-only, and Both are mutually
exclusive error counts.
Union is the percentage of all errors
captured by either method; Jaccard measures
overlap between the two captured-error sets.
Macro means are unweighted across 12 settings.
}
\label{tab:complementarity_overlap}
\end{table*}

Table~\ref{tab:complementarity_auroc} and Figure \ref{fig:complementarity} (top) show that
the hybrid outperforms B-CARE in nine settings,
with the three exceptions occurring on SemEval.
The hybrid achieves the highest AUROC among
the three approaches in eight settings.
Table~\ref{tab:complementarity_overlap} and Figure \ref{fig:complementarity} (bottom)
show a mean captured-error Jaccard overlap
of 0.406 and a mean union capture rate of
55.5\%. Together, these findings suggest
that the two methods identify partially
distinct errors, although their combination
does not consistently improve ranking
performance.

\section{Component Ablation: Detailed Results}
\label{app:score_ablation}

Table~\ref{tab:component_ablation_full} reports
the AUROC of the full B-CARE score ($A-S$),
support-only ($-S$), and alternative-only ($A$)
across all 12 dataset--model settings.
We also report paired AUROC differences and
95\% confidence intervals based on 10,000
paired bootstrap resamples. Intervals are
unadjusted for multiple comparisons.

\begin{table*}[t]
\centering
\footnotesize
\setlength{\tabcolsep}{4.2pt}
\renewcommand{\arraystretch}{1.08}

\begin{tabular}{llccccccl}
\toprule
\textbf{Dataset} &
\textbf{Model} &
\textbf{B-CARE} &
\textbf{$-S$} &
\textbf{$A$} &
\textbf{$\Delta_{-S}$} &
\textbf{$\Delta_A$} &
\textbf{95\% CI ($\Delta_{-S}$)} &
\textbf{95\% CI ($\Delta_A$)} \\
\midrule

\multirow{3}{*}{SemEval}
& Phi
& \textbf{0.7969} & 0.7823 & 0.7933
& +0.0146 & +0.0036
& [0.0093, 0.0203] & [-0.0058, 0.0131] \\

& Mistral
& 0.8490 & 0.8372 & \textbf{0.8521}
& +0.0119 & -0.0031
& [0.0081, 0.0160] & [-0.0096, 0.0036] \\

& Gemma
& 0.7378 & 0.7191 & \textbf{0.7419}
& +0.0187 & -0.0041
& [0.0119, 0.0262] & [-0.0143, 0.0065] \\

\midrule

\multirow{3}{*}{Vaccines}
& Phi
& 0.6984 & \textbf{0.7011} & 0.6162
& -0.0026 & +0.0822
& [-0.0103, 0.0052] & [0.0637, 0.1014] \\

& Mistral
& \textbf{0.6723} & 0.6711 & 0.6093
& +0.0013 & +0.0630
& [-0.0059, 0.0088] & [0.0452, 0.0809] \\

& Gemma
& \textbf{0.7243} & 0.7228 & 0.6427
& +0.0015 & +0.0815
& [-0.0066, 0.0095] & [0.0632, 0.0999] \\

\midrule

\multirow{3}{*}{COVID}
& Phi
& \textbf{0.6903} & 0.6772 & 0.6692
& +0.0131 & +0.0211
& [0.0045, 0.0225] & [0.0051, 0.0376] \\

& Mistral
& \textbf{0.6840} & 0.6800 & 0.6635
& +0.0039 & +0.0205
& [-0.0030, 0.0115] & [0.0060, 0.0358] \\

& Gemma
& \textbf{0.7068} & 0.6971 & 0.6795
& +0.0097 & +0.0273
& [0.0020, 0.0181] & [0.0118, 0.0438] \\

\midrule

\multirow{3}{*}{P-Stance}
& Phi
& \textbf{0.7616} & 0.7610 & 0.6591
& +0.0007 & +0.1026
& [-0.0027, 0.0042] & [0.0800, 0.1254] \\

& Mistral
& \textbf{0.7655} & 0.7606 & 0.6780
& +0.0049 & +0.0876
& [0.0010, 0.0090] & [0.0656, 0.1097] \\

& Gemma
& \textbf{0.7282} & 0.7258 & 0.6290
& +0.0024 & +0.0992
& [-0.0018, 0.0068] & [0.0736, 0.1246] \\

\midrule

\multicolumn{2}{l}{\textbf{Macro mean}}
& \textbf{0.7346} & 0.7279 & 0.6861
& +0.0067 & +0.0484
& --- & --- \\

\bottomrule
\end{tabular}

\caption{
Detailed component ablation across 12
dataset--model settings (AUROC).
$\Delta_{-S}$ and $\Delta_A$ denote the
AUROC differences between B-CARE and the
support-only and alternative-only variants,
respectively. Confidence intervals are based
on 10,000 paired bootstrap resamples and
are unadjusted for multiple comparisons.
Bold indicates the highest AUROC per setting.
}
\label{tab:component_ablation_full}
\end{table*}

\section{Graded Evidence versus Evidence-Label Disagreement}
\label{app:graded_binary_ablation}

We compare B-CARE with Evidence-Label
Disagreement (LD), a binary baseline using
identical frozen, prediction-blind verifier
outputs, without additional LLM calls.
For base prediction $\hat{y}_B$ and ordinal classwise
evidence $E_c$, the two risk scores are
\begin{equation}
\begin{aligned}
R_B &= \max_{c\neq\hat{y}_B} E_c-E_{\hat{y}_B},\\
R_{\mathrm{LD}} &= \mathbb{I}\!\left[E_{\hat{y}_B}\leq
\max_{c\neq\hat{y}_B}E_c
\right].
\end{aligned}
\end{equation}
Thus, LD flags disagreement unless the base
prediction has uniquely maximal evidence; ties
are treated as disagreement.

\paragraph{AUROC comparison.}
Table~\ref{tab:binary_full_auroc} reports results
across four datasets and three base models.
B-CARE outperforms LD in all 12 settings, with
all unadjusted 95\% paired-bootstrap confidence
intervals excluding zero. The macro-average
AUROC increases from approximately 0.6543
to 0.7346, corresponding to a mean improvement
of approximately 0.0803.

\begin{table*}[t]
\centering
\small
\setlength{\tabcolsep}{6pt}
\begin{tabular}{llcccc}
\toprule
\textbf{Dataset} & \textbf{Base} &
\textbf{B-CARE} & \textbf{LD} &
$\boldsymbol{\Delta}$ &
\textbf{95\% CI} \\
\midrule
\multirow{3}{*}{SemEval}
& Phi     & 0.7969 & 0.7346 & +0.0622 & [0.0424, 0.0805] \\
& Mistral & 0.8490 & 0.8028 & +0.0462 & [0.0313, 0.0605] \\
& Gemma   & 0.7378 & 0.6816 & +0.0562 & [0.0339, 0.0776] \\
\midrule
\multirow{3}{*}{Vaccines}
& Phi     & 0.6984 & 0.5883 & +0.1102 & [0.0867, 0.1344] \\
& Mistral & 0.6723 & 0.5901 & +0.0822 & [0.0610, 0.1034] \\
& Gemma   & 0.7243 & 0.6133 & +0.1110 & [0.0886, 0.1335] \\
\midrule
\multirow{3}{*}{COVID}
& Phi     & 0.6903 & 0.5990 & +0.0913 & [0.0637, 0.1199] \\
& Mistral & 0.6840 & 0.6166 & +0.0674 & [0.0423, 0.0950] \\
& Gemma   & 0.7068 & 0.6150 & +0.0918 & [0.0657, 0.1181] \\
\midrule
\multirow{3}{*}{P-Stance}
& Phi     & 0.7616 & 0.6782 & +0.0834 & [0.0625, 0.1051] \\
& Mistral & 0.7655 & 0.6889 & +0.0766 & [0.0560, 0.0969] \\
& Gemma   & 0.7282 & 0.6437 & +0.0845 & [0.0603, 0.1084] \\
\midrule
\multicolumn{2}{l}{\textbf{Macro average}}
& \textbf{0.7346} & 0.6543 &
\textbf{+0.0803} & --- \\
\bottomrule
\end{tabular}
\caption{AUROC comparison of graded B-CARE and
Evidence-Label Disagreement across 12 dataset--model settings.
$\Delta$ denotes B-CARE minus LD. Confidence
intervals are unadjusted 95\% paired-bootstrap
intervals (10,000 resamples).}
\label{tab:binary_full_auroc}
\end{table*}

\begin{figure}[t]
\centering
\includegraphics[width=\columnwidth]
{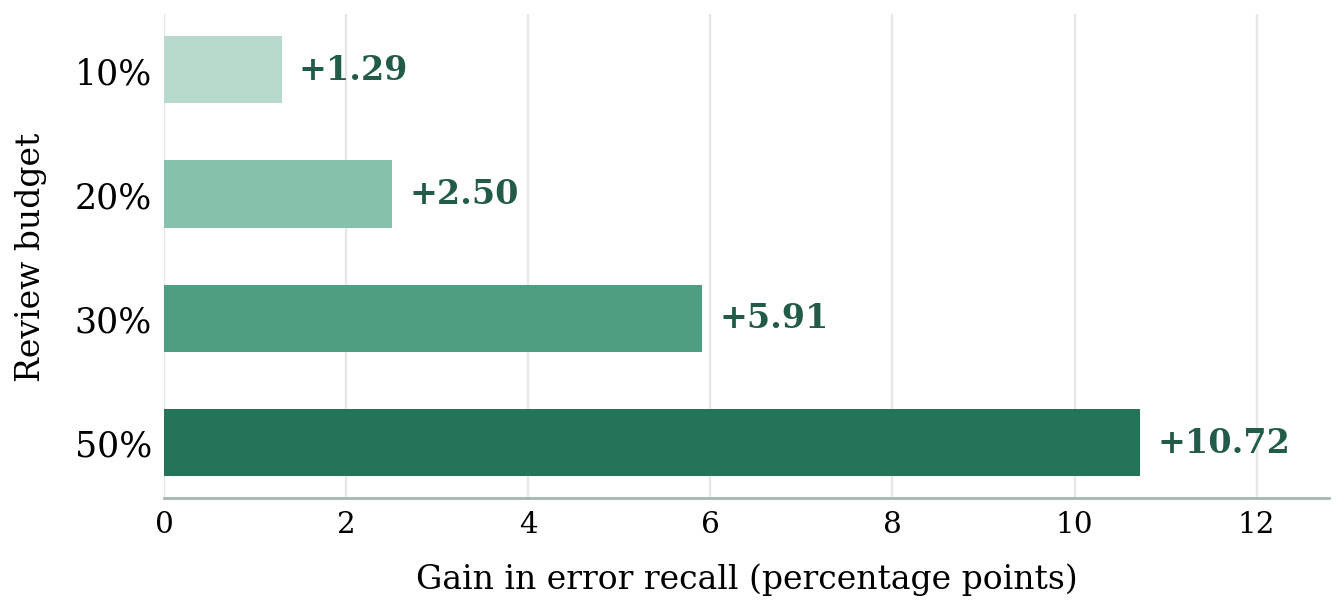}
\caption{Macro-average improvement in expected
error recall (percentage points) for B-CARE
over Evidence-Label Disagreement at fixed review budgets.
Results are averaged equally across 12 settings:
SemEval, Vaccines, COVID-19, and P-Stance, each
evaluated with Phi, Mistral, and Gemma.
At a 50\% review budget, mean recall is
0.7426 for B-CARE versus 0.6354 for LD;
this comparison is descriptive.}
\label{fig:binary_review_budget}
\end{figure}

\paragraph{Precision--recall and review budgets.}
The advantage extends beyond AUROC
(Table~\ref{tab:binary_additional}; Figure \ref{fig:binary_review_budget}).
B-CARE improves AUPRC in all 12 settings and
achieves higher mean expected error recall at
10\%, 20\%, and 30\% review budgets.
Expected recall accounts for random tie-breaking
within equal-risk groups.

\begin{table}[!t]
\centering
\small
\setlength{\tabcolsep}{5pt}
\begin{tabular}{lccc}
\toprule
\textbf{Metric} &
$\boldsymbol{\Delta}$ &
\textbf{Wins} &
\textbf{CI $>0$} \\
\midrule
AUROC       & +0.0803 & 12/12 & 12/12 \\
AUPRC       & +0.0813 & 12/12 & 12/12 \\
Recall@10\% & +0.0129 & 12/12 & 6/12 \\
Recall@20\% & +0.0250 & 11/12 & 5/12 \\
Recall@30\% & +0.0591 & 12/12 & 12/12 \\
\bottomrule
\end{tabular}
\caption{Macro-average gains of B-CARE over LD.
Wins count settings with positive point-estimate
differences; CI $>0$ counts unadjusted 95\%
paired-bootstrap intervals excluding zero
in favor of B-CARE.}
\label{tab:binary_additional}
\end{table}

\paragraph{Within-group ranking and tie sensitivity.}
B-CARE retains ranking information even among
instances assigned the same LD score: its mean
within-group AUROC is 0.6469 for LD$=0$ and
0.5764 for LD$=1$. Furthermore, changing the
tie policy to count a prediction as agreement
whenever it belongs to the maximal-evidence
set does not alter the conclusion: B-CARE
outperforms LD in all 12 settings under both
tie conventions.

% ============================================================
% B-CARE: Matched RunPod Evaluation — Revised Appendix
% Required packages: graphicx, booktabs, amsmath
% Figures are stored in figures/
% ============================================================

\section{Matched Self-Hosted Verifier Evaluation}
\label{app:runpod_matched}

\paragraph{Motivation and experimental setting.}
We conducted a supplementary matched comparison
between B-CARE and LlamaProb, a probability-based
verifier comparator, using a self-hosted
Llama-3.3-70B-Instruct model on RunPod.
Dedicated GPU resources were used because
availability and scheduling contention on the
university's shared computing infrastructure
constrained timely execution. This decision
was operational rather than motivated by
benchmark performance.

Both scoring methods were evaluated using the
same local verifier deployment, the same test
instances, and identical frozen base-model
predictions. The local Hugging Face checkpoint
was Llama-3.3-70B-Instruct, with a revision
identifier beginning \texttt{6f6073b4}.
The frozen base models were Phi-4,
Mistral-Small-24B-Instruct-2501, and
Gemma-3-27B.

This matched evaluation is distinct from the
primary OpenRouter-based experiments.
In particular, the RunPod verifier evidence
and the original OpenRouter evidence should
not be treated as interchangeable.
The comparison is intended to examine
alternative error-ranking signals under
a common local inference environment,
rather than differences between
inference providers.

\paragraph{Error-risk definitions.}
For an input text $x$, stance target $t$,
and frozen base-model prediction $\hat{y}_B$,
B-CARE obtains prediction-blind classwise
evidence scores
$E_c(x,t)\in\{0,1,2,3\}$.

Its error-risk score is

\begin{equation}
R_{\mathrm{B}}(x,t;\hat{y}_B)
=
\max_{c\neq\hat{y}_B} E_c(x,t)
-
E_{\hat{y}_B}(x,t).
\label{eq:runpod_bcare_risk}
\end{equation}

The resulting margin takes seven possible
integer values, from $-3$ to $+3$.

LlamaProb instead computes

\begin{equation}
R_{\mathrm{LP}}(x,t;\hat{y}_B)
=
1-p_V(\hat{y}_B\mid x,t),
\label{eq:runpod_lp_risk}
\end{equation}

where $p_V$ denotes the verifier's
closed-set probability assigned to the
frozen base prediction.

In the audited RunPod implementation,
LlamaProb probabilities were obtained
using teacher-forced label log-probabilities
and FP32 softmax normalization.
Gold labels were not used to construct
either risk score.

Higher scores indicate greater estimated
error risk for both methods.

Although both methods use the same verifier,
they differ in their elicitation and scoring
procedures. Accordingly, this comparison
evaluates two complete error-ranking
approaches rather than isolating the
effect of score discretization alone.

\begin{figure}[t]
\centering
\includegraphics[width=\columnwidth]
{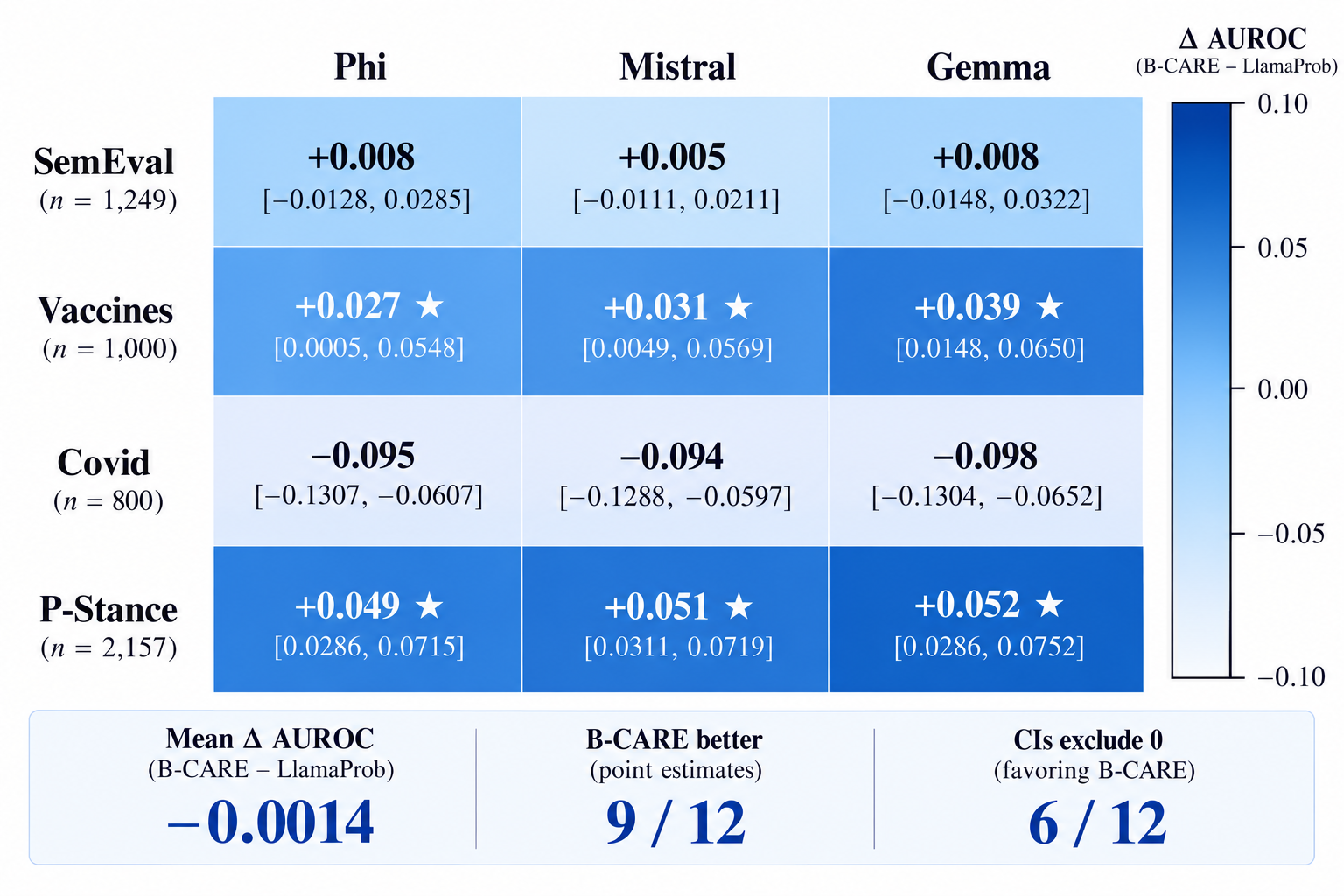}

\caption{
AUROC differences between B-CARE and
LlamaProb across four datasets and three
frozen base models. Positive differences
favor B-CARE; negative differences favor
LlamaProb. Blue intensity represents
the absolute magnitude of the difference,
not statistical significance.
}
\label{fig:runpod_delta_auroc}
\end{figure}

\paragraph{Matched AUROC comparison.}
Table~\ref{tab:runpod_matched_auroc} and Figure \ref{fig:runpod_delta_auroc}
presents the AUROC results across four
datasets and three frozen base models.

B-CARE achieves higher AUROC in nine
of the twelve settings, covering all
three base models on SemEval, Vaccines,
and P-Stance. The improvements range from $0.005$ to
$0.008$ on SemEval, $0.027$ to $0.039$
on Vaccines, and $0.049$ to $0.052$
on P-Stance. COVID exhibits the opposite pattern:
LlamaProb achieves higher AUROC across
all three base models.

Consequently, despite nine setting-level
wins for B-CARE, the unweighted mean
AUROC across the twelve settings is
approximately $0.760$ for B-CARE
and $0.761$ for LlamaProb.
The number of setting-level wins is
descriptive and does not itself
constitute a statistical significance
test.

% ============================================================
% TABLE 1: MATCHED AUROC
% ============================================================

\begin{table}[t]
\centering
\begingroup
\setlength{\tabcolsep}{3.3pt}
\renewcommand{\arraystretch}{1.07}

\resizebox{\columnwidth}{!}{%
\begin{tabular}{llrrr}
\toprule
\textbf{Dataset} &
\textbf{Model} &
\textbf{B-CARE} &
\textbf{LP} &
$\boldsymbol{\Delta}$ \\
\midrule
\multirow{3}{*}{SemEval}

& Phi & \textbf{0.794} & 0.786 & +0.008 \\
& Mistral & \textbf{0.845} & 0.840 & +0.005 \\
& Gemma & \textbf{0.736} & 0.728 & +0.008 \\

\midrule
\multirow{3}{*}{Vaccines}

& Phi & \textbf{0.719} & 0.692 & +0.027 \\
& Mistral & \textbf{0.702} & 0.671 & +0.031 \\
& Gemma & \textbf{0.757} & 0.718 & +0.039 \\

\midrule
\multirow{3}{*}{COVID}

& Phi & 0.720 & \textbf{0.815} & -0.095 \\
& Mistral & 0.710 & \textbf{0.804} & -0.094 \\
& Gemma & 0.729 & \textbf{0.827} & -0.098 \\

\midrule
\multirow{3}{*}{P-Stance}

& Phi & \textbf{0.810} & 0.761 & +0.049 \\
& Mistral & \textbf{0.819} & 0.768 & +0.051 \\
& Gemma & \textbf{0.780} & 0.728 & +0.052 \\

\midrule

Mean (12 settings)
& & 0.760 & \textbf{0.761} & -0.001 \\

\bottomrule
\end{tabular}%
}
\endgroup

\caption{
Matched RunPod AUROC comparison between
B-CARE and LlamaProb (LP).
$\Delta=\mathrm{B\text{-}CARE}-\mathrm{LP}$;
positive values favor B-CARE.
Best AUROC values are shown in bold.
Values are displayed to three decimal places.
}
\label{tab:runpod_matched_auroc}
\end{table}

% ============================================================
% FIGURE 1: AUROC HEATMAP
% ============================================================

% ============================================================
% COVID DIAGNOSTICS
% ============================================================

\paragraph{Diagnosing the COVID performance reversal.}
To investigate LlamaProb's consistent advantage on COVID,
we analyzed matched per-instance outputs for 800 test
examples under each of the three frozen base models.
Table~\ref{tab:runpod_covid_diagnostics} reports the
AUROC differences, paired bootstrap confidence intervals,
and the contribution of B-CARE score ties.

LlamaProb outperforms B-CARE by approximately
0.094--0.097 AUROC across the three models.
The 95\% confidence intervals, estimated using 10,000
paired bootstrap resamples, exclude zero in all three
settings. An implementation audit independently
recomputed the AUROCs and found no discrepancies in
the error labels, evaluation procedure, or LlamaProb
probability extraction.

\begin{table}[t]
\centering
\begingroup
\setlength{\tabcolsep}{4pt}
\renewcommand{\arraystretch}{1.12}
\resizebox{\columnwidth}{!}{%
\begin{tabular}{lccc}
\toprule
\textbf{Model} &
$\boldsymbol{\Delta}$\textbf{AUROC} &
\textbf{95\% CI} &
\textbf{Tie share} \\
\midrule
Phi
& -0.095
& $[-0.131,-0.061]$
& 72.5\% \\
Mistral
& -0.094
& $[-0.129,-0.060]$
& 68.3\% \\
Gemma
& -0.097
& $[-0.130,-0.065]$
& 66.5\% \\
\bottomrule
\end{tabular}%
}
\endgroup
\caption{
COVID diagnostic results from the matched RunPod
evaluation ($n=800$).
$\Delta$AUROC is B-CARE minus LlamaProb.
Confidence intervals are estimated from 10,000 paired
bootstrap resamples.
Tie share is the percentage of the observed AUROC
gap attributable to error--correct pairs tied by B-CARE.
The Gemma difference is calculated from unrounded
scores and may differ from subtraction of displayed
three-decimal AUROCs.
}
\label{tab:runpod_covid_diagnostics}
\end{table}

\paragraph{Ordinal compression and tied rankings.}
A substantial portion of the observed COVID
performance gap is associated with the limited
ranking resolution of B-CARE's ordinal evidence
margin.
Because classwise evidence scores belong to
$\{0,1,2,3\}$, the resulting risk margin has only
seven possible values.

A substantial proportion of COVID instances receives
the minimum risk score, $R_{\mathrm{B}}=-3$:
51.4\% for Phi and Mistral and 48.3\% for Gemma.
In contrast, LlamaProb produces approximately 790
distinct scores among the 800 COVID examples.

Consequently, approximately 21--24\% of
error--correct instance pairs are tied under B-CARE.
These pairs cannot be distinguished by its final
risk margin, even when their underlying texts or
verifier evidence profiles differ.

\paragraph{Pairwise AUROC decomposition.}
AUROC can be expressed as the fraction of
error--correct pairs correctly ordered by risk,
with tied pairs receiving half credit:

\begin{equation}
\mathrm{AUROC}
=
\frac{
N_{\mathrm{win}}
+\frac{1}{2}N_{\mathrm{tie}}
}{
N_{+}N_{-}
},
\label{eq:runpod_auroc_pairwise}
\end{equation}

where $N_{+}$ and $N_{-}$ denote the numbers of
incorrect and correct base-model predictions,
respectively.

For pairs tied under B-CARE, its pairwise
AUROC contribution is necessarily $0.5$.
LlamaProb, however, can distinguish these pairs,
achieving a conditional pairwise AUROC of
approximately 0.78--0.80.

Let $N_{\mathrm{B,tie}}$ denote the number of
error--correct pairs tied by B-CARE, and let
$A_{\mathrm{LP}\mid\mathrm{B,tie}}$ denote
LlamaProb's conditional pairwise AUROC on
those pairs.
Their contribution to the B-CARE-minus-LlamaProb
AUROC difference is

\begin{equation}
\Delta_{\mathrm{tie}}
=
\left(
\frac{1}{2}
-
A_{\mathrm{LP}\mid\mathrm{B,tie}}
\right)
\frac{
N_{\mathrm{B,tie}}
}{
N_{+}N_{-}
}.
\label{eq:runpod_tie_decomposition}
\end{equation}

This decomposition isolates the observed ranking
difference among pairs that B-CARE assigns equal
risk scores.
Such pairs account for 72.5\%, 68.3\%, and 66.5\%
of the total AUROC gap for Phi, Mistral, and Gemma,
respectively.

Thus, approximately two-thirds to three-quarters
of LlamaProb's COVID advantage arises from
comparisons that B-CARE cannot resolve.

As an additional diagnostic, preserving B-CARE's
ordering between distinct risk scores while using
LlamaProb to break ties increases AUROC to
approximately 0.77--0.79.
This hybrid ranking is used solely to quantify
within-bin information loss; it is not proposed
as a new prediction-blind method.

\paragraph{Class-conditional and target-specific effects.}
The remaining COVID differences are not explained
primarily by the NONE class.
Removing instances with gold or predicted NONE
labels increases LlamaProb's relative advantage,
whereas B-CARE performs slightly better on the
gold-NONE subset.

A substantial portion of the residual disagreement
occurs among instances predicted as AGAINST.
Within this predicted class, AUROC differences
frequently favor LlamaProb by approximately
0.14--0.16.

One recurring pattern involves incorrect AGAINST
predictions for which the verifier nevertheless
assigns strong ordinal evidence to AGAINST.
The resulting B-CARE margin can be $-2$ or $-3$,
indicating low estimated error risk.
For some of these instances, LlamaProb assigns
near-zero probability to the predicted label,
producing a substantially higher error-risk score.

The disagreement is especially pronounced for
the Face Masks target, where LlamaProb achieves
AUROC of approximately 0.93--0.94 and its
advantage ranges from approximately 0.11 to 0.19.

However, this pattern is not uniform across
COVID targets.
School Closures provides a counterexample:
both methods exhibit relatively weak
discrimination, and B-CARE achieves a higher
AUROC point estimate.

These findings indicate that the aggregate
COVID reversal reflects both ordinal score
compression and target-specific ranking
disagreements rather than uniform failure
across all targets or stance classes.

\paragraph{Interpretation of the COVID diagnostics.}
The pairwise decomposition identifies score ties
as the largest measured contributor to the
COVID AUROC gap.
Nevertheless, this accounting is descriptive:
it does not establish that discretization alone
causes the entire difference between the methods.

Likewise, the observed AGAINST disagreement
does not establish why the verifier assigns
particular evidence scores without additional
instance-level annotation.

The confidence intervals and tie decomposition
are specific to the matched RunPod diagnostic
evaluation. Their numerical values should not
be assumed to transfer directly to the original
OpenRouter-generated evidence.

\paragraph{Error capture under limited review budgets.}
AUROC measures discrimination across all
error--correct pairs, whereas practical
error auditing often involves reviewing only
a limited number of high-risk instances.

\begin{figure}[t]
\centering
\includegraphics[width=\columnwidth]
{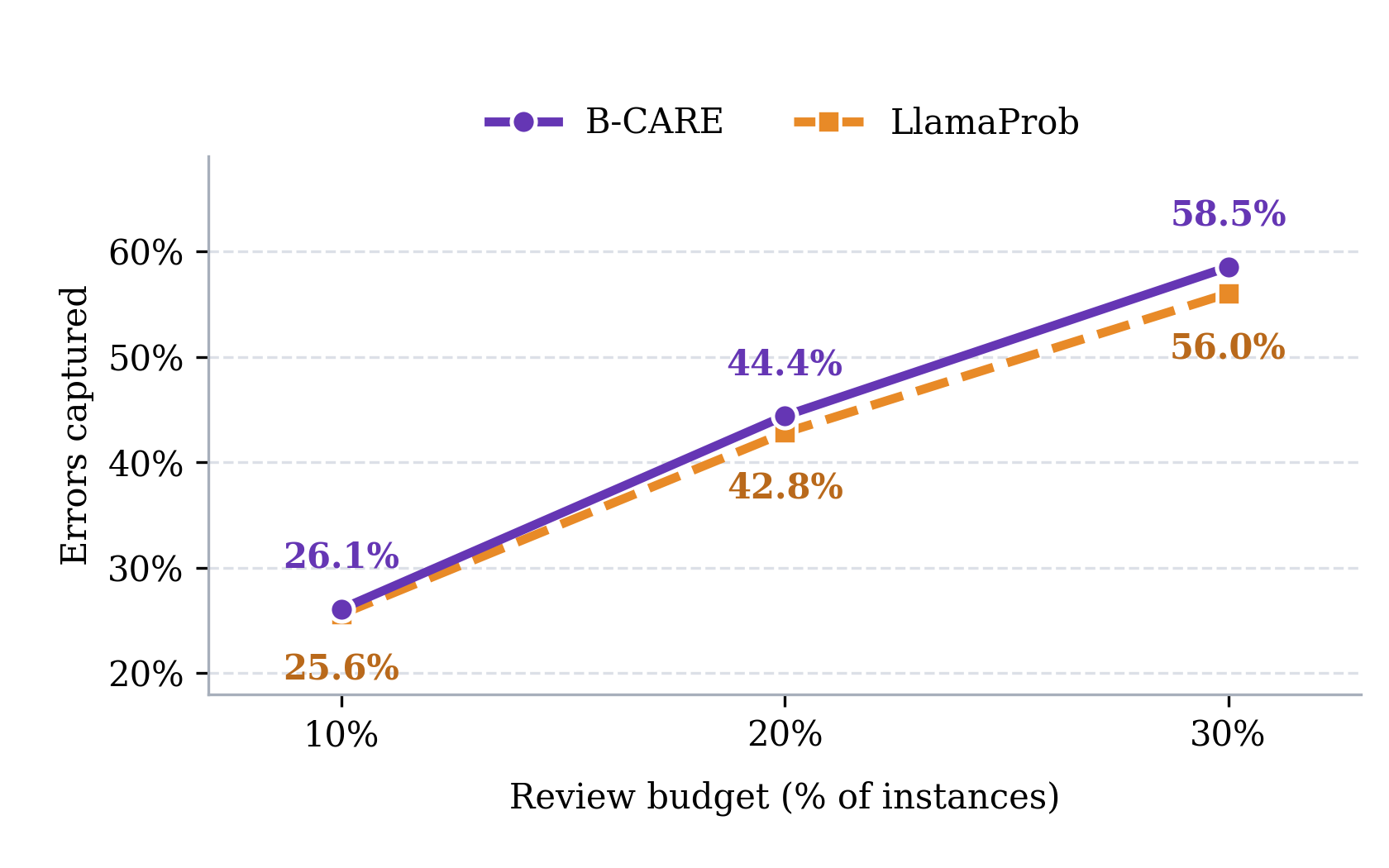}
\caption{
Mean error capture across twelve matched
dataset--model settings at review budgets
of 10\%, 20\%, and 30\%.
B-CARE is shown in purple and LlamaProb
in orange. Higher values indicate better
error capture.
}
\label{fig:runpod_error_capture}
\end{figure}

Figure~\ref{fig:runpod_error_capture}
compares the mean fraction of errors retrieved
at review budgets of 10\%, 20\%, and 30\%
across the twelve matched dataset--model settings.

B-CARE captures 26.1\%, 44.4\%, and 58.5\%
of errors at these budgets, compared with
25.6\%, 42.8\%, and 56.0\% for LlamaProb.
These correspond to absolute improvements of
0.5, 1.6, and 2.5 percentage points,
respectively.

These aggregate differences do not imply
that B-CARE performs better at every budget
or within every dataset--model setting.

\begin{table}[t]
\centering
\begingroup
\setlength{\tabcolsep}{5pt}
\renewcommand{\arraystretch}{1.1}
\begin{tabular}{lcc}
\toprule
\textbf{Metric} &
\textbf{B-CARE} &
\textbf{LlamaProb} \\
\midrule
Mean AUROC $\uparrow$
& 0.760 & \textbf{0.761} \\
Mean AUPRC $\uparrow$
& \textbf{0.603} & 0.592 \\
Mean E-AURC $\downarrow$
& 0.116 & \textbf{0.108} \\
EC@10\% $\uparrow$
& \textbf{26.1\%} & 25.6\% \\
EC@20\% $\uparrow$
& \textbf{44.4\%} & 42.8\% \\
EC@30\% $\uparrow$
& \textbf{58.5\%} & 56.0\% \\
\bottomrule
\end{tabular}
\endgroup
\caption{
Aggregate matched RunPod metrics averaged
across twelve dataset--model settings.
Arrows indicate the preferred direction.
EC denotes error capture at the specified
review budget.
}
\label{tab:runpod_aggregate}
\end{table}

\paragraph{Comparison across ranking criteria.}
Table~\ref{tab:runpod_aggregate} summarizes
the aggregate performance of the two methods
across complementary ranking metrics.

B-CARE achieves higher mean AUPRC
(0.603 versus 0.592) and higher mean
error capture at all three evaluated budgets.
LlamaProb achieves a marginally higher
mean AUROC (0.761 versus 0.760) and
lower mean E-AURC (0.108 versus 0.116).

These results are compatible because the
metrics emphasize different ranking properties.
AUROC measures global error--correct pairwise
discrimination, whereas AUPRC is sensitive
to precision--recall behavior and error
prevalence. Error capture evaluates retrieval
within a specified review budget, while
E-AURC summarizes performance across
the risk--coverage curve.

Accordingly, improved error capture at
selected budgets does not necessarily
imply superior global AUROC.

The matched results show that B-CARE's
prediction-blind ordinal evidence can support
competitive error ranking and early error
retrieval, while LlamaProb's continuous
scores provide additional discrimination
within some B-CARE risk bins.
Neither method consistently dominates
across all reported criteria.

\end{document}